\documentclass[twocolumn, switch]{article} 

\usepackage{preprint}

\usepackage{amsmath, amsthm, amssymb, amsfonts}

\usepackage[numbers,square]{natbib}
\usepackage[utf8]{inputenc}	
\usepackage[T1]{fontenc}	
\usepackage{xcolor}		
\usepackage[colorlinks = true,
            linkcolor = purple,
            urlcolor  = blue,
            citecolor = cyan,
            anchorcolor = black]{hyperref}	
\usepackage{booktabs} 		
\usepackage{nicefrac}		
\usepackage{microtype}		
\usepackage{lineno}		
\usepackage{float}			

\usepackage{lipsum}		

\usepackage{newfloat}
\DeclareFloatingEnvironment[name={Supplementary Figure}]{suppfigure}
\usepackage{sidecap}
\sidecaptionvpos{figure}{c}

\usepackage{titlesec}
\titlespacing\section{0pt}{12pt plus 3pt minus 3pt}{1pt plus 1pt minus 1pt}
\titlespacing\subsection{0pt}{10pt plus 3pt minus 3pt}{1pt plus 1pt minus 1pt}
\titlespacing\subsubsection{0pt}{8pt plus 3pt minus 3pt}{1pt plus 1pt minus 1pt}

\usepackage{tikz,xcolor,hyperref}

\definecolor{lime}{HTML}{A6CE39}
\DeclareRobustCommand{\orcidicon}{
	\begin{tikzpicture}
	\draw[lime, fill=lime] (0,0) 
	circle [radius=0.16] 
	node[white] {{\fontfamily{qag}\selectfont \tiny ID}};
	\draw[white, fill=white] (-0.0625,0.095) 
	circle [radius=0.007];
	\end{tikzpicture}
	\hspace{-2mm}
}
\foreach \x in {A, ..., Z}{\expandafter\xdef\csname orcid\x\endcsname{\noexpand\href{https://orcid.org/\csname orcidauthor\x\endcsname}
			{\noexpand\orcidicon}}
}

\title{The Potential of Visual ChatGPT For Remote Sensing}

\usepackage{authblk}

\author[1\thanks{\tt{lucasosco@unoeste.br}}]{Lucas Prado Osco\orcidA{}}
\author[2]{Eduardo Lopes de Lemos\orcidB{}}
\author[2]{Wesley Nunes Gonçalves\orcidC{}}
\author[3]{Ana Paula Marques Ramos\orcidD{}}
\author[4]{José Marcato Junior\orcidE{}}

\affil[1]{\scriptsize Faculty of Engineering and Architecture and Urbanism, University of Western São Paulo (UNOESTE), Rod. Raposo Tavares, km 572, Limoeiro, Presidente Prudente 19067-175, SP, Brazil; lucasosco@unoeste.br; pradoosco@gmail.com}
\affil[2]{Faculty of Computing, Federal University of Mato Grosso do Sul (UFMS), Av. Costa e Silva-Pioneiros, Cidade Universitária, Campo Grande 79070-900, MS, Brazil; lopes.eduardo@ufms.br, wesley.goncalves@ufms.br}
\affil[3]{Departament of Cartography, São Paulo State University (UNESP), Centro Educacional, R. Roberto Simonsen, 305, Presidente Prudente, 19060-900, SP, Brazil; marques.ramos@unesp.br}
\affil[4]{Faculty of Engineering, Architecture and Urbanism and Geography, Federal University of Mato Grosso do Sul (UFMS), Av. Costa e Silva-Pioneiros, Cidade Universitária, Campo Grande 79070-900, MS, Brazil; jose.marcato@ufms.br}

\begin{document}

\twocolumn[ 
  \begin{@twocolumnfalse} 
  
\maketitle

\begin{abstract}
\small Recent advancements in Natural Language Processing (NLP), particularly in Large Language Models (LLMs), associated with deep learning-based computer vision techniques, have shown substantial potential for automating a variety of tasks. These are known as Visual LLMs and one notable model is Visual ChatGPT, which combines ChatGPT's LLM capabilities with visual computation to enable effective image analysis. These models' abilities to process images based on textual inputs can revolutionize diverse fields, and while their application in the remote sensing domain remains unexplored, it is important to acknowledge that novel implementations are to be expected into it. Thus, this is the first paper to examine the potential of Visual ChatGPT, a cutting-edge LLM founded on the GPT architecture, to tackle the aspects of image processing related to the remote sensing domain. Among its current capabilities, Visual ChatGPT can generate textual descriptions of images, perform canny edge and straight line detection, and conduct image segmentation. These offer valuable insights into image content and facilitate the interpretation and extraction of information. By exploring the applicability of these techniques within publicly available datasets of satellite images, we demonstrate the current model's limitations in dealing with remote sensing images, highlighting its challenges and future prospects. Although still in early development, we believe that the combination of LLMs and visual models holds a significant potential to transform remote sensing image processing, creating accessible and practical application opportunities in the field.
\end{abstract}
\vspace{0.35cm}

  \end{@twocolumnfalse} 
] 



\section{Introduction}

Remote sensing image processing is a critical task for monitoring and analyzing the Earth's surface and environment. It is used in a wide range of fields such as agriculture, forestry, geology, water resources, and urban planning \cite{Yuan2020, Osco2021}. However, analyzing and interpreting large volumes of remote sensing data can be time-consuming and labor-intensive, requiring specialized knowledge and expertise \cite{Osco2021}. In recent years, Large Language Models (LLMs) emerged as powerful and innovative tools for human assistance in various domains \cite{ge2023openagi}, holding the potential to be implemented in the remote sensing area as well. 

As Artificial Intelligence (AI) continues to evolve, novel models demonstrate an unprecedented ability to understand and generate human-like text, as well as perform numerous tasks based on human guidance \cite{zhao2023survey}. Among the LLMs, a model named ChatGPT stands out as a remarkable example, offering immense promise for assisting humans in multiple activities. The Generative Pre-trained Transformer (GPT), a deep learning model developed by OpenAI \cite{openai2023gpt4}, has gained considerable attention as a promising AI technique for natural language processing tasks. This VLM not only consists in one of the most recent foundation model in development, but as well one of the prominent in its field since has gained notoriety by the public eye in recent times.

The GPT model has been trained on extensive text data and can generate human-like responses to input prompts. This model is particularly useful in tasks such as chatbots, text summarization, and language translation \cite{openai2023gpt4, liu2023summary}. Recent research, however, has explored the application of LLMs models in visual tasks such as image generation, captioning, and analysis assistance \cite{zhang2023adding}. These models, also known as Visual Language Models (VLMs), can generate natural language descriptions of images and perform image processing tasks from text descriptions. One model that is gaining attention is the Visual ChatGPT \cite{wu2023visual}. Visual ChatGPT is an extension of ChatGPT that incorporates visual information on its capabilities while also providing text-based responses in a conversational style.

Although still in its early concepts, the fusion of LLMs and visual models may revolutionize image processing and unlock new practical applications in various fields \cite{zhang2023visionlanguage}. In this context, remote sensing is an area that could directly benefit from this integration. Fine-tuned VLMs could potentially be used to process and analyze satellite and aerial images to detect land use changes, monitor natural disasters, and assess environmental impacts, as well as assist in the classification and segmentation of images for easier interpretation and decision-making.

In this paper, we discuss the significance, utility, and limitations of the model Visual ChatGPT in assisting humans in remote sensing image processing. This model has shown great potential in various applications such as question-answering systems and image generation and modification. Currently, Visual ChatGPT can perform image processing tasks like edge detection, line extraction, and image segmentation, which are interesting for the remote sensing field. The model, however, is not fine-tuned to deal with the remote sensing domain, thus making it still an early adoption of the tool. Regardless, we investigate this, as a basis for discussion of its potential, by comparing these tools within publicly available datasets of remote sensing imagery, thus measuring its capabilities both quantitatively and qualitatively.

By enabling machines to understand and generate images, Visual ChatGPT paves the way for numerous applications in image processing. Herein, we discussed how Visual ChatGPT can be adapted to the remote-sensing domain, where it might revolutionize the way we process and analyze these images. We examined state-of-the-art developments in the model, evaluated their capabilities in the context of remote sensing imagery, and proposed future research directions. Ultimately, this exploration seeks to provide insights into the integration of VLMs into remote sensing science and community.

\section{Visual ChatGPT: A Revolution in Image Analysis and its Potential in Remote Sensing}

Visual ChatGPT is an advanced VLM that combines the capabilities of text-based LLMs with visual understanding. This revolutionary approach enables machines to analyze images and generate relevant text or visual outputs, opening up new possibilities for image analysis and processing. One of the key features of Visual ChatGPT is its ability to incorporate state-of-the-art algorithms and information into its current model, facilitating continuous improvement and adaptation \cite{wu2023visual}. 

By fine-tuning the model with domain-specific datasets, Visual ChatGPT can become increasingly proficient in specific tasks, making it an invaluable tool for image analysis. With its architecture built to process and analyze both textual and visual information, it has the potential to revolutionize diverse fields. Interaction with Visual ChatGPT involves a dynamic and iterative process, where users can provide textual input, image data, or both, and the model responds with relevant information or actions. This flexibility allows for a wide range of tasks to be performed, including generating images from the user input text, providing photo descriptions, answering questions about images, performing object and pose detection, as well as other various image processing techniques, such as edge detection, straight line detection, scene classification, and image segmentation, which are interesting in the remote sensing context. 

Image processing methods are essential for extracting valuable information from remote sensing data. However, these techniques often require additional computational knowledge and can be challenging for non-specialists to implement. VLMs like Visual ChatGPT offer the potential to bridge this knowledge gap by providing an accessible interface for non-experts to analyze image data. 

Although still early in its conception, many techniques and methods can be integrated into VLMs, thus providing the means to perform complex image processing \cite{zhang2023adding, zhang2023visionlanguage}. In remote sensing, tasks such as edge and line detection, scene classification, and image segmentation, which currently are some of the techniques embedded into Visual ChatGPT’s model, can be used to perform and enhance the analysis of aerial or satellite imagery and bring important information to the end user.

Edge detection is an image processing technique that identifies the boundaries between different regions or objects within an image. In remote sensing, edge detection is vital for recognizing features on the Earth's surface, such as roads, rivers, and buildings, and others \cite{abraham2021edge}. Visual ChatGPT, with its ability to analyze images and generate relevant text or visual outputs, can be adapted to assist non-experts in executing edge detection tasks of different objects present in the image. By providing textual input alongside image data, users can interact with the model to identify boundaries and extract valuable information about the scene being analyzed.

Straight line detection is another critical image processing technique in remote sensing, with applications in feature extraction. It involves identifying linear targets in remote sensing images, such as roads, rivers, and boundaries \cite{Kumar2020}. Visual ChatGPT can be utilized to help non-experts perform line detection tasks by processing image data and easily returning line pattern identification in the images. This capability enables users to extract additional information about the underlying terrain or land use and cover without requiring in-depth knowledge of these image-processing techniques.

Scene classification and image segmentation are also essential techniques in remote sensing for identifying different types of land cover and separating them into distinct regions. These techniques aid in monitoring land use changes, detecting deforestation, assessing urban growth, monitoring water reservoirs, and estimating agriculture growth, among many others \cite{Kotaridis2021}. On this, VLMs can be employed to facilitate scene classification and image segmentation tasks for non-experts. In scene classification, Visual ChatGPT can be used to detect and describe objects in the image. As for segmentation, with specifically fine-tuned models, there is the potential for users to obtain results by simply interacting with the model using textual input \cite{li2023transformerbased}, allowing them to analyze land changes and monitor impacts.

However, it is important to note that the current version of Visual ChatGPT has not been yet specifically trained on remote sensing imagery. Neither have any other VLMs precisely tuned for this task since the technology is still in an early stage. Nonetheless, the model's architecture and capabilities offer a solid foundation for fine-tuning and adapting it to this domain in future implementations. 

By training Visual ChatGPT on remote sensing datasets, it is possible that it can be tailored to recognize and analyze unique features, patterns, and structures present in aerial or satellite images. To fully realize its potential, thorough analysis and evaluation of its usage, impact, practices, and errors in remote sensing applications are necessary. This will not only assist the development of improved VLMs but also pave the way for more efficient, accurate, and comprehensive analyses of remote sensing data performed by these tools.

\section{Materials and Methods}

\begin{figure*}[ht!]
\centering
\includegraphics[width=\textwidth]{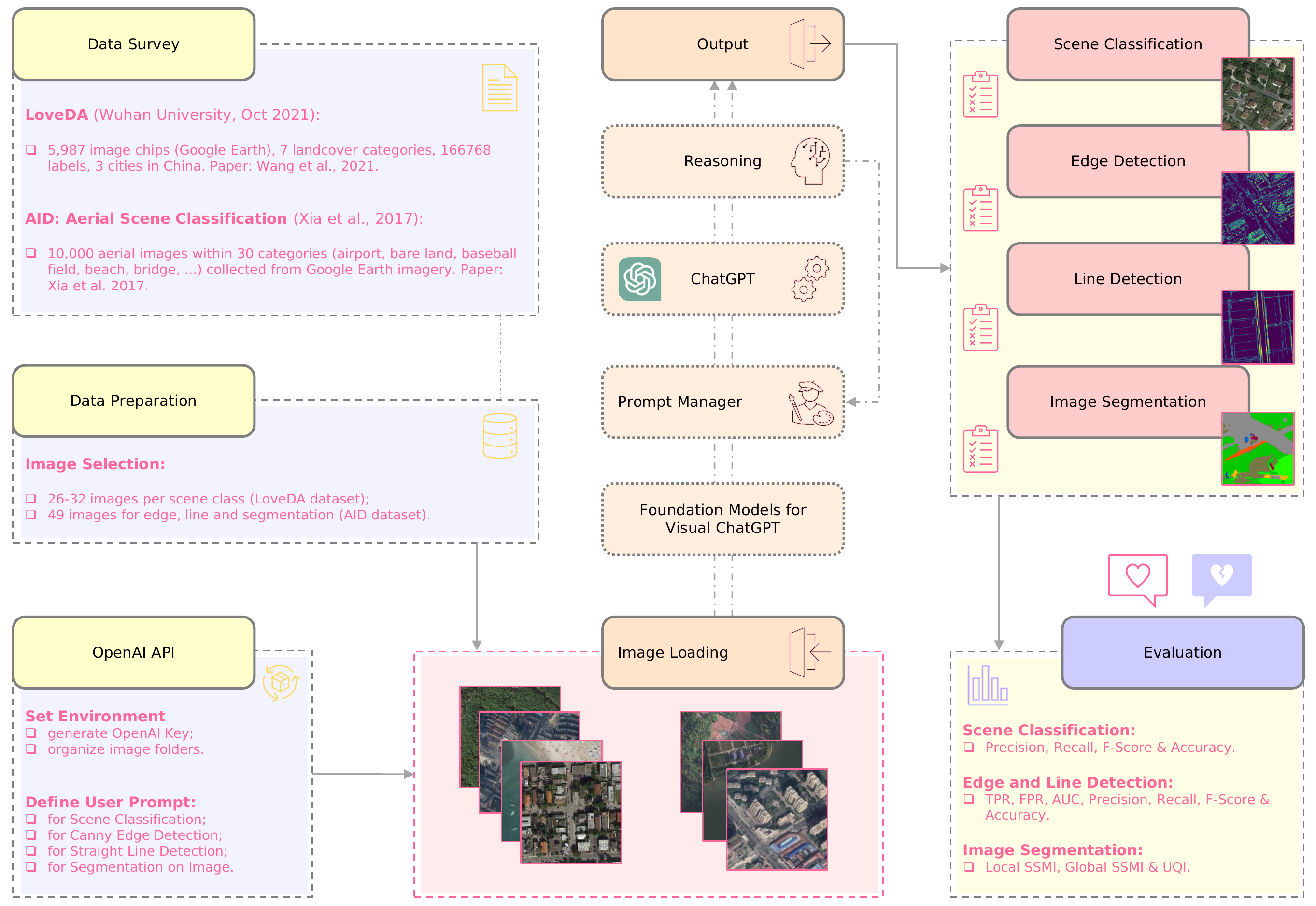}
\caption{\small \centering Diagram of the evaluation process of Visual ChatGPT in remote sensing image processing tasks. The diagram follows an up-down/left-to-right flow, indicating that the process begins with a data survey, preparation, and setting up of the environment for loading the images into Visual ChatGPT. Next, different tasks are performed using the tools provided by Visual ChatGPT, and the results are stored for analysis where different sets of metrics are applied to evaluate the performance of the model.\label{fig_method}}
\end{figure*}

In this section, we detail the materials and methods used to evaluate the performance of Visual ChatGPT in remote sensing image processing tasks. The evaluation process is divided into several stages (Figure~\ref{fig_method}), focusing on different aspects of the models' current capabilities, mainly on image classification, edge, and straight line detection, and image segmentation.

We initiated our evaluation of Visual ChatGPT by assessing its performance in scene classification tasks. To this end, we used a publicly available dataset containing Google Earth images labeled by human specialists. We extracted a small portion of this dataset, considering a subset of its classes for our tests. The model's classification performance was compared to the ground-truth labels provided in the dataset.

In the next stage, we qualitatively evaluated the edge and straight line detection capabilities of Visual ChatGPT on remote sensing imagery, from Google Earth, of another publicly available dataset. The detected edges and lines were assessed to determine the model's effectiveness in identifying target features in the images. The model's performance was compared with traditional edge filters and manually labeled lines.

Lastly, we evaluated the image segmentation feature of Visual ChatGPT using the images from the same previous dataset, which was specifically designed for segmentation data training. We then compared the resulting segmentations with their corresponding masks. The comparison was conducted using an associative method in which the classes identified by the Visual ChatGPT model were associated with the classes labeled in the dataset.

\subsection{Experiment Delineation}

To implement Visual ChatGPT, we downloaded the code from Microsoft Github \cite{microsoft_taskmatrix}, created a virtual environment, installed the required dependencies, downloaded the pre-trained models, and started a Flask server. Once the server was running, we imported the required libraries on Python code and set the API key for the OpenAI platform access. The “run\_image” function inside the original “visual\_chatgpt.py” file was modified to handle image resizing and captioning. Next, the Visual ChatGPT model was loaded with the required sub-models. 

It is important to point out that Visual ChatGPT provides a different set of tools, but not all of them are appropriate to deal with tasks related to remote sensing images. In this sense, we used only the following: “Get Photo Description”, “Answer Question About The Image”, “Edge Detection On Image”, “Line Detection On Image” and “Segmentation On Image”. Our code then loops through a folder containing the images and performs the canny edge and straight line detection, as well as segmentation on each image. It also obtains the default image description of the original loaded image using the Visual ChatGPT model and then asks a classification question to determine the class of the image. The results are then stored in a .csv file and used for further evaluation.

Visual ChatGPT utilizes sub-models that are specifically designed to cater to the different prompts and tools required. For instance, the "Get Photo Description" and "Answer Question About The Image" tools use models from the HuggingFace library \cite{li2022blip} to generate natural language descriptions of an image and answer questions based on the given image path and the corresponding question. The "Edge Detection On Image" tool uses the Canny Edge Detector \cite{canny1986computational} from the OpenCV library to identify and detect the edges of an image when given its path. Similarly, the "Line Detection On Image" tool uses the M-LSD Detector for Straight Line model \cite{gu2022lightweight} to detect straight lines in the image. Finally, the "Segmentation On Image" tool employs the UniFormer Segmentation model \cite{li2022uniformer} to segment different classes on the given image.

To assess the effectiveness of the Visual Chat-GPT models in handling remote sensing image data, we surveyed publicly available datasets related to this field. After consideration, we selected two datasets that would allow us to investigate the model's capabilities for performing specific tasks. These datasets were the "AID: Aerial Scene Classification" \cite{Xia2017} and the "LoveDA: A Remote Sensing Land-Cover Dataset for Domain Adaptive Semantic Segmentation" \cite{wang2022loveda}. Both datasets contain Google Earth imagery captured at different times, with varying lighting conditions and visualization scales. These datasets provide a rich and diverse set of images that are well-suited for testing the model's performances. 

In its current form, the computational cost of using Visual ChatGPT is slightly higher than traditional methods. This increased cost primarily stems from the necessity of consuming tokens within the OpenAI API. The tokens required to process each input and produce the corresponding output can add up, particularly in large-scale image-processing tasks. As technology and computational efficiency evolve, we anticipate a reduction in these costs in the near future. However, at the moment, this cost influenced the number of runs conducted throughout our experiment, as we detail in the description of each dataset.

The AID dataset contains different scene classes with about 200 to 400 samples of 600x600 size for each class, with 10,000 images in total. However, due to the current cost associated with using Visual ChatGPT, we randomly selected between 26 to 32 images of each class for evaluation. These images were reviewed to ensure that a broad representation of possible inputs were selected. The following classes were evaluated: “Airport”, “BareLand”, “BaseballField”, “Beach”, “Bridge”, “Center”, “Church”, “Commercial”, “DenseResidential”, “Desert”, “Farmland”, “Forest”, “Industrial”, “Meadow”, “MediumResidential”, “Mountain”, “Park”. These were stored in a “classes” variable within our code. We chose these 17 classes to ensure a diverse representation of the scenes, since the remaining classes provided similar context. This brought a total of 515 images to be loaded and described (and, therefore, classified) by the Visual ChatGPT model. These images were used for evaluating the “Get Photo Description”, and “Answer Question About The Image” tools.

The LoveDA dataset is composed of 5,987 image chips, being segmented into 7 landcover categories (namely: "background", "building", "road", "water", "barren", "forest" and "farmland), totaling 166,768 labels across 3 cities. This dataset focuses on multi-geographical environments, variating between “Urban” and “Rural” characteristics, while providing challenges like multi-scale objects presence; complex background samples, and inconsistent class distributions. The dataset also provides the segmentation masks used to train image models. Here we used these masks as our “ground-truth” data and selected a small portion of the dataset, consisting of 49 images (mixing both “Urban” and “Rural” environments). These 49 image chips were all used in the evaluation of the “Edge Detection On Image”, “Line Detection On Image” and “Segmentation On Image” tools. They represent the most complex and rich environments within their respective geographical context, and were limited due to the cost associated with the API's usage.

As mentioned, for the latter, we utilized a purposive sampling methodology to directly select remote sensing images representative of different land covers. Our objective was to maintain a rich representation of diverse surface covers in our dataset. As such, to ensure a comprehensive depiction of geographical scenarios, we, in this case, directly hand-picked images that provided views of both natural and man-made environments. This approach is grounded in the intention to not just create a representative dataset but to ensure that our dataset reflects the complexities and variances that are inherently present in real-world scenarios. In doing so, we believe that the chosen dataset yielded more robust and generalized outcomes in subsequent analyses and applications.

\subsection{Protocol for Scene Classification Evaluation}

We first investigated whether Visual ChatGPT can assist in classifying remote sensing scenes. To test this, we used the AID dataset (Aerial Scene Classification) \cite{Xia2017}. We evaluated the "Get Photo Description" and “Answer Question About The Image” functions of Visual ChatGPT by asking it to describe and classify the selected images. For each image, we asked Visual ChatGPT to choose, based on its image description, with which class it would associate the image. We directly asked it to choose between each one of the 17 classes, instead of trying to guess them, thus generating guided predictions. A file was created with the stored results and compared the Visual ChatGPT classification with the correct class from the dataset.

We used the confusion matrix to evaluate the performance of Visual ChatGPT in classifying the scenes. The confusion matrix is a commonly used tool in the evaluation of classification models. It provides a summary of the performance of a model by showing the number of correct and incorrect predictions for each class. We begin by loading the dataset into a data frame. The set contains two columns, “Image” and “Answer to the Question”, that correspond to the true and predicted labels for each data point, respectively.

The classes were defined as a list of strings representing the different categories in the dataset. The two mentioned columns were then converted and used for generating the confusion matrix. The matrix takes as input the true labels (y\_true), predicted labels (y\_pred), and the list of class labels (classes). Finally, a heatmap was created to represent it. The heatmap was customized by adding annotations to show the number of predictions in each cell. We calculated the Precision, Recall, F-Score and Accuracy metrics to assess the performance of Visual ChatGPT in comparison to the correct class labeled from the AID dataset. These metrics can be described as follows \cite{powers2020evaluation}:\\

\noindent Precision: Precision measures the proportion of True Positive (TP) instances among the instances that were predicted as positive. Higher precision means fewer False Positives (FP).

\begin{linenomath}
\begin{equation}
\text{Precision} = \frac{\text{TP}}{(\text{TP} + \text{FP)}}
\end{equation}
\end{linenomath}

\noindent Recall: Recall measures the proportion of TP instances among the actual positive instances, thus using False Negatives (FN) into its equation. This metric works better when considering binary tasks.

\begin{linenomath}
\begin{equation}
\text{Recall} = \frac{TP}{(TP + FN)}
\end{equation}
\end{linenomath}

\noindent F-Score: F-Score is the harmonic mean of Precision and Recall. It's a balanced metric that considers both false positives and false negatives, with a range from 0 (worst) to 1 (best).

\begin{linenomath}
\begin{equation}
\text{F Score} = 2 * \frac{(Precision * Recall)}{(Precision + Recall)}
\end{equation}
\end{linenomath}

\noindent Overall Accuracy: Accuracy is the proportion of correct predictions (both TP and TN) among the total number of instances. While it's a commonly used metric, it is not suitable for imbalanced datasets.

\begin{linenomath}
\begin{equation}
\text{Accuracy} = \frac{(TP + TN)}{(TP + FP + TN + FN)}
\end{equation}
\end{linenomath}

Taking into account the substantial number of classes in this problem (n=17), we computed the baseline accuracy to provide a context for evaluating the model's overall performance. The baseline accuracy, also referred to as "random chance," signifies the probability of accurately identifying a class by merely selecting the most prevalent class, as:

\begin{linenomath}
\begin{equation}
\text{Baseline Accuracy} = \max_i \frac{N_i}{N_\text{total}}
\end{equation}
\end{linenomath}
where:\\
\`i` represents each class in the dataset\\
\`N\_i` is the number of images in class `i`\\
\`N\_\text{total}` is the total number of images in the dataset.\\

\subsection{Protocol for Edge and Line Detection Evaluation}

For the edge and line detections, we asked Visual ChatGPT to perform both the “Edge Detection On Image”, and “Line Detection On Image” functions, extracting the edge and straight line features in the images. To investigate its capabilities, we compared them with two traditional edge detection methods, the Canny filter \cite{canny1986computational} and the Sobel filter \cite{Sobel1990AnI3} , and with manual annotation of straight lines present in the images. Both filters were manually fine-tuned over the same images to provide the overall most interesting results, thus differentiating from the default, fully-automated approach, of Visual ChatGPT. For this, we used the selected 49 images from the LoveDa dataset \cite{wang2022loveda} to be processed by the filters and compared. The Python programming language was utilized for this implementation, relying on the NumPy, imageio, and scikit-image libraries.

First, the image file was loaded where a function was employed to read the image in grayscale format, simplifying the image for further processing. The resulting image matrix was converted into a floating-point data type and normalized to the range of [0, 1] by dividing each pixel value by 255. This normalization step was crucial for maintaining consistency across images and ensuring the edge detection algorithms could process them appropriately.

The Canny edge detection filter was applied to the normalized grayscale images. This was accomplished by passing the image and a sigma value, varying between 1 and 3, to its function. The sigma parameter determines the amount of Gaussian smoothing applied to the image, effectively controlling the sensitivity of the algorithm to any noise. The Canny edge detection filter aims to identify continuous edges in an image by performing non-maximum suppression and double thresholding to remove unwanted pixels \cite{canny1986computational}. The resulting edge map consists of pixels representing the detected edges. 

Next, the Sobel edge detection filter was applied to the normalized grayscale images by implementing its  function. This calculates the gradient magnitude at each pixel in the image, and the output is a continuous-valued edge map, providing an approximation of the edge intensity \cite{Sobel1990AnI3}. The Sobel edge detection algorithm is a simpler method. It is based on the convolution of the image with two 3x3 kernels, one for the horizontal gradient and one for the vertical gradient. This method is computationally efficient and straightforward but may be more susceptible to noise compared to the Canny edge detection filter. 

After applying both edge detection filters, we saved the resulting images as 8-bit grayscale images into separate folders. The conversion to 8-bit grayscale format was performed by multiplying the processed image arrays by 255 and then casting them to the unsigned 8-bit integer data type  before saving them. The data was stored to later be used to compare against the edge detection performed by Visual ChatGPT.

For the straight line detection approach, we compare the results of the straight lines detected by Visual ChatGPT with manually labeled lines from the dataset. The manually labeled lines served as the ground-truth for evaluating its performance. For this, we identified, in the same 49 images, line aspects like roads, rivers, plantations, and terrain that resembled linear characteristics and that are of overall interest when dealing with remote sensing data. These images were saved and stored in a folder to be promptly loaded and compared.

As such, we compared both the line and edge detection performances following the same protocol. To achieve this, we defined a function to load and preprocess the images. This function takes two image file paths as input (one from Visual ChatGPT and the other from our “ground-truth”) and performs the following steps: 1. Load the images in the grayscale format; 2. Resize both images to the same dimensions (512x512 pixels); 3. Apply Otsu's thresholding method to obtain the optimal threshold for each image to create edge and line binary maps, and; 4. Flatten the binary maps into 1D arrays for extracting the comparison metrics.

Finally, for each image pair, we called the process\_images function to obtain the performance metrics and stored them in a list called “results”. After processing the images, we calculated various performance metrics, such as True Positive Rate (TPR), False Positive Rate (FPR), Area Under the Curve (AUC), as well as Precision, Recall, F-Score, and Accuracy using scikit-learn's metrics module. These metrics were essential for evaluating and comparing the performance of the methods in terms of their ability to identify true and false lines and edges, and overall accuracy. Since we already explained Precision, Recall, F-Score, and Accuracy, the remaining metrics to be described are \cite{powers2020evaluation}:\\

\noindent True Positive Rate (TPR): TPR is the proportion of TP instances among the actual positive instances. The higher the TPR, the better the model is at identifying true lines and edges.

\begin{linenomath}
\begin{equation}
\text{TPR} = \frac{TP}{(TP + FN)}
\end{equation}
\end{linenomath}

\noindent False Positive Rate (FPR): FPR is the proportion of FP instances among the True Negative (TN) instances. The lower the FPR, the better the model is at avoiding false edge and line detections.

\begin{linenomath}
\begin{equation}
\text{FPR} = \frac{FP}{(FP + TN)}
\end{equation}
\end{linenomath}

\noindent Area Under the Curve (AUC): AUC is a measure of the overall performance of a classification model. It's calculated by plotting the Receiver Operating Characteristic (ROC) curve, which shows the trade-off between TPR and FPR. AUC ranges from 0 to 1, where a higher value indicates better performance.

\subsection{Protocol for Image Segmentation Evaluation}

To evaluate the performance of Visual ChatGPT's image segmentation capabilities on remote sensing data, we used the previously separated 49 images from the LoveDa dataset \cite{wang2022loveda}, which includes manually labeled data as masks to segmentation training. The protocol used for this task comprises a two-step procedure by comparing the Visual ChatGPT’s segmented output with the manually labeled ground-truth images. This VLM uses the "Segmentation on Image" function, which brings the Unified transFormer (UniFormer) \cite{li2022uniformer} model to perform image segmentation.

The Unified transFormer (UniFormer) is a model developed to handle both local redundancy and complex global dependency typically found in visual data. This model blends the merits of Convolution Neural Networks (CNNs) and Vision Transformers (ViTs) in a unified format. UniFormer incorporates three crucial modules: Dynamic Position Embedding (DPE), Multi-Head Relation Aggregator (MHRA), and Feed-Forward Network (FFN). DPE, as an initial step, dynamically incorporates position information into all tokens, which is particularly effective for visual recognition with arbitrary input resolution. Next, MHRA enhances each token by exploring its contextual tokens through relation learning. MHRA fuses convolution and self-attention, mitigating local redundancy while capturing global dependencies. Lastly, FFN enhances each token individually, following the typical ViTs approach, encompassing two linear layers and a non-linear function (GELU) \cite{li2022uniformer}.

Since Visual ChatGPT doesn’t know which classes to look at on the image, it tries to guess them based on its current capabilities when implementing the “Segmentation on Image” function. Thus, it is not possible to perform a "direct" comparison between the ground-truth classes with which the class Visual ChatGPT imagines it to be. Therefore, metrics like Precision, Recall, F-Score, and Accuracy are not feasible to evaluate this task. Since we are comparing two segmented images with different classes, we opted to use metrics that quantify the similarity or dissimilarity between the images and determine how well they align with each other. To achieve this, we extracted two key metrics: the Structural Similarity Index Measure (SSIM) \cite{Wang2004} and the Universal Image Quality Index (UQI) \cite{ZhouWang2002}.

The SSIM is a metric used to measure the similarity between two images or patches based on structural information. It ranges between -1 and 1, with 1 indicating a perfect match and -1 indicating a complete mismatch. The Sewar library likely provides local and global SSIM values. Local SSIM averages the score, providing a fine-grained evaluation and identifying local variations in image quality. Global SSIM computes the score for the entire image, providing a holistic evaluation of overall similarity. Having both local and global SSIM scores can help identify areas or regions where image quality is poorer or the modifications have had a more significant impact. The SSIM equations (both Local and Global) are defined by \cite{Wang2004}:

\begin{linenomath}
\begin{equation}
\text{SSIM(x, y)} = \frac{(2\mu_x\mu_y + C_1)(2\sigma_{xy} + C_2)}{(\mu_x^2 + \mu_y^2 + C_1)(\sigma_x^2 + \sigma_y^2 + C_2)}
\end{equation}
\end{linenomath}
where:\\
\(x\) and \(y\) are local regions (patches) of the two images being compared\\
\(\mu_x\) and \(\mu_y\) are the average intensities of the patches \(x\) and \(y\)\\
\(\sigma_x^2\) and \(\sigma_y^2\) are the variances of the patches \(x\) and \(y\)\\
\(\sigma_{xy}\) is the covariance between the patches \(x\) and \(y\)\\
\(C_1\) and \(C_2\) are small constants to stabilize the division (typically, \(C_1 = (K_1L)^2\) and \(C_2 = (K_2L)^2\), where \(L\) is the dynamic range of the pixel values, and \(K_1\) and \(K_2\) are small constants)\\

\begin{linenomath}
\begin{equation}
\text{Global\ SSIM(X, Y)} = \frac{1}{N} \sum_{i=1}^{N} SSIM(x_i, y_i)
\end{equation}
\end{linenomath}
where:\\
\(X\) and \(Y\) are the two images being compared\\
\(x_i\) and \(y_i\) are local patches of the images \(X\) and \(Y\)\\
\(N\) is the number of local patches in the images\\

The UQI is a full-reference image quality metric that compares processed images with the original or reference image (ground-truth in this case). It measures the similarity between images using their structural information, based on their luminance and contrast. The UQI calculates the mean, standard deviation, and covariance of luminance and contrast values for the two images, and combines them using a weighted average to obtain a final UQI value ranging from 0 to 1. Thus, higher UQI values indicate higher image quality and similarity between the processed and reference images. This metric is widely used to evaluate image processing and compression algorithms for both objective and subjective image quality evaluations. The UQI is defined by the following equation \cite{ZhouWang2002}:

\begin{linenomath}
\begin{equation}
\text{UQI(X, Y)} = \frac{4\sigma_{XY}\mu_X\mu_Y}{(\sigma_X^2+\sigma_Y^2)(\mu_X^2+\mu_Y^2)}
\end{equation}
\end{linenomath}
where:\\
\(X\) and \(Y\) are the two images being compared\\
\(\mu_X\) and \(\mu_Y\) are the average intensities of the images \(X\) and \(Y\)\\
\(\sigma_X^2\) and \(\sigma_Y^2\) are the variances of the images \(X\) and \(Y\)\\
\(\sigma_{XY}\) is the covariance between the images \(X\) and \(Y\)\\

In the first part of the procedure, we preprocessed the ground-truth images. We begin by loading the black and white images and converting them to grayscale using the PIL library. Then, a color map was defined, assigning a specific color to each of the 7-pixel values in the ground-truth image. These colors were defined based on the colors used by Visual ChatGPT to return segmented regions of similar characteristics. By iterating over the width and height of each image, the black and white images were converted to colored images using this color map. The final step involves resizing the colored image to a 512x512 resolution and saving it to the appropriate directory.

The second part of the procedure focuses on computing the image quality metrics. To accomplish this, the necessary libraries were imported, including the Sewar library for full-reference image quality metrics, the imageio library for image input/output, and the skimage library for image processing. We then defined a list of dictionaries containing the file paths for pairs of the ground-truth and the predicted images. As the function iterates through each image pair, it loads, normalizes, and resizes the ground-truth and predicted images to the desired size of 512x512 pixels. The images are then converted back to uint8 format. For each image pair, we calculate the SSIM and UQI metrics using the Sewar library. These metrics were stored in a dictionary and appended to a list.

The SSIM and UQI metrics served as valuable tools for assessing the performance of Visual ChatGPT's image segmentation, considering our current limitation on dealing with different classes. In summary, these metrics were chosen because the SSIM measures the structural similarity between the predicted and ground-truth images, taking into account changes in similarity and structures, while the UQI provides a scalar value indicating the overall quality of the predicted image in comparison to the ground-truth image. By analyzing these metrics, it was possible to identify areas where the segmentation model excels or falters, assisting in guiding further model improvement and evaluation.

\section{Results}

\subsection{Scene Classification}

\begin{figure*}[h!]
\centering
\includegraphics[width=12.5cm]{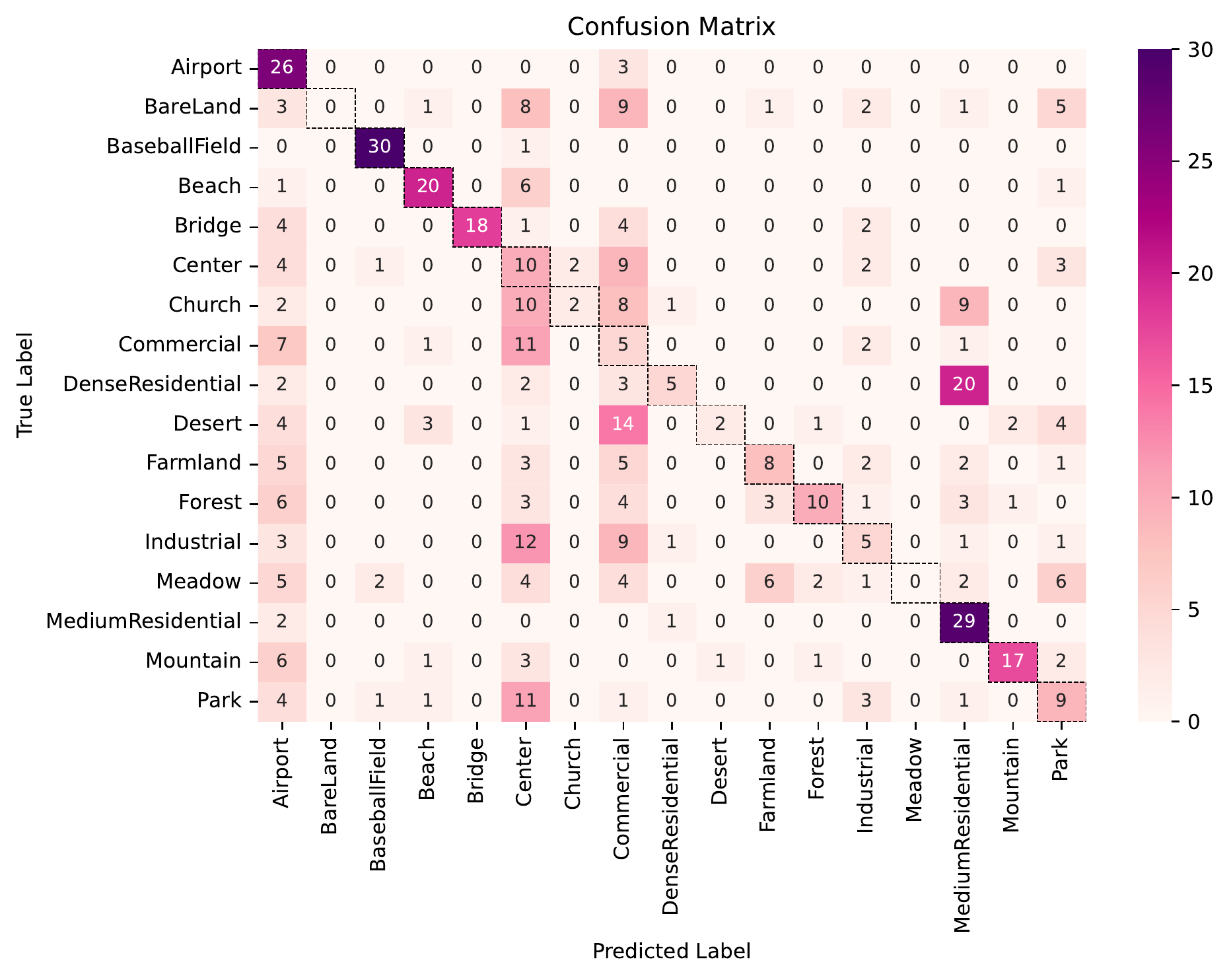}
\caption{\small \centering Confusion matrix from the evaluated portion of the AID dataset classified by Visual ChatGPT. The color intensity and the numeric values within each cell of the heatmap indicate the number of instances of the predicted label.\label{fig_scene1}}
\end{figure*}

We initially evaluated Visual ChatGPT's ability to classify remote sensing scenes using the AID dataset \cite{Xia2017}. To support this analysis, Figure~\ref{fig_scene1} presents a heatmap visualization of the calculated confusion matrix, generated from the scene classification predictions. 

\begin{figure*}[h!]
\centering
\includegraphics[width=12cm]{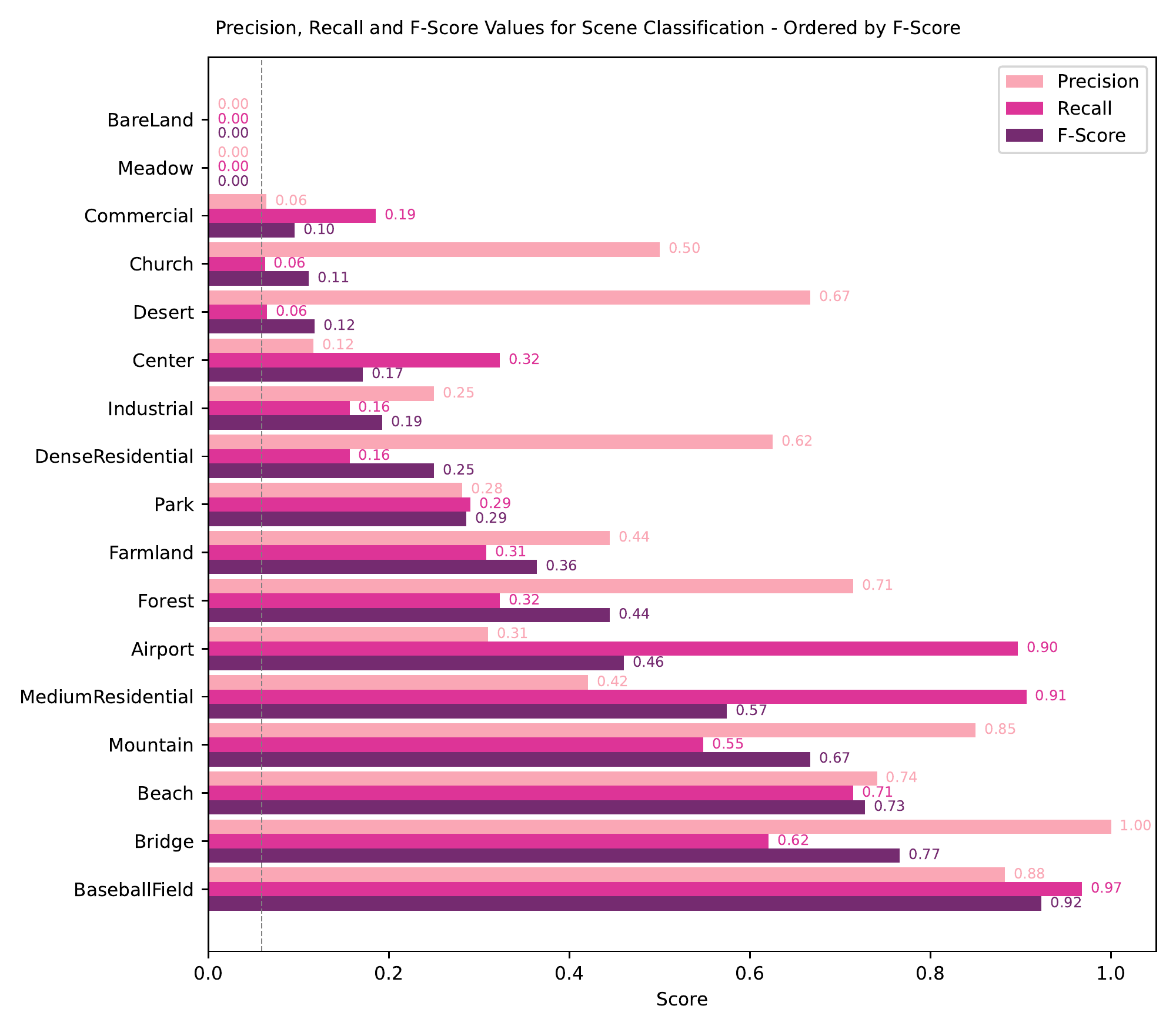}
\caption{\small \centering Evaluation metrics from the AID dataset image classified by Visual ChatGPT. The Precision, Recall, and F-Score values are displayed, sorted by F-Score from lowest to highest. A grey dashed vertical line is plotted at a score of 0.0588, serving as a visual reference point for comparison, indicating the "random-chance" point.\label{fig_scene2}}
\end{figure*}

Based on the confusion matrix, we also calculated the Precision, Recall, and F-Score metrics and displayed them in a horizontal bar chart, presented in Figure~\ref{fig_scene2}. The overall accuracy of the model for this task was 0.381 (or 38.1\%), with the averaged weighted values between all the classes as 0.583 (58.3\%), 0.381 (38.1\%), and 0.359 (35.9\%) for Precision, Recall, and F-Score, respectively.

The selected classes offered valuable insights into the model's ability to interpret satellite imagery. The graphics (Figures~\ref{fig_scene1} and~\ref{fig_scene2}) demonstrated that the model more accurately identified scenes containing Baseball Fields, Bridges, Beaches, and Mountains, as evidenced by the high F-Scores achieved. Conversely, it struggled to recognize landscapes such as Bareland, Meadows, and Deserts, resulting in lower performance metrics. Additionally, the model encountered difficulties in distinguishing urban scenes, including Commercial, Church, Center, Industrial, and Dense Residential areas. This was indicated by high Precision values, but low Recall and F-Scores, which fell significantly below the "random-guess" threshold.

Although the overall accuracy of the model is 38.1\%, which might seem relatively low, it's important to consider the context of the problem with 17 classes. The "random chance" (baseline accuracy) for this classification task is about 5.88\%. Furthermore, the Visual ChatGPT model effectively interpreted and classified a considerable number of images across various classes, demonstrating its potential for handling remote sensing imagery. 

\begin{figure*}[h!]
\centering
\includegraphics[width=17cm]{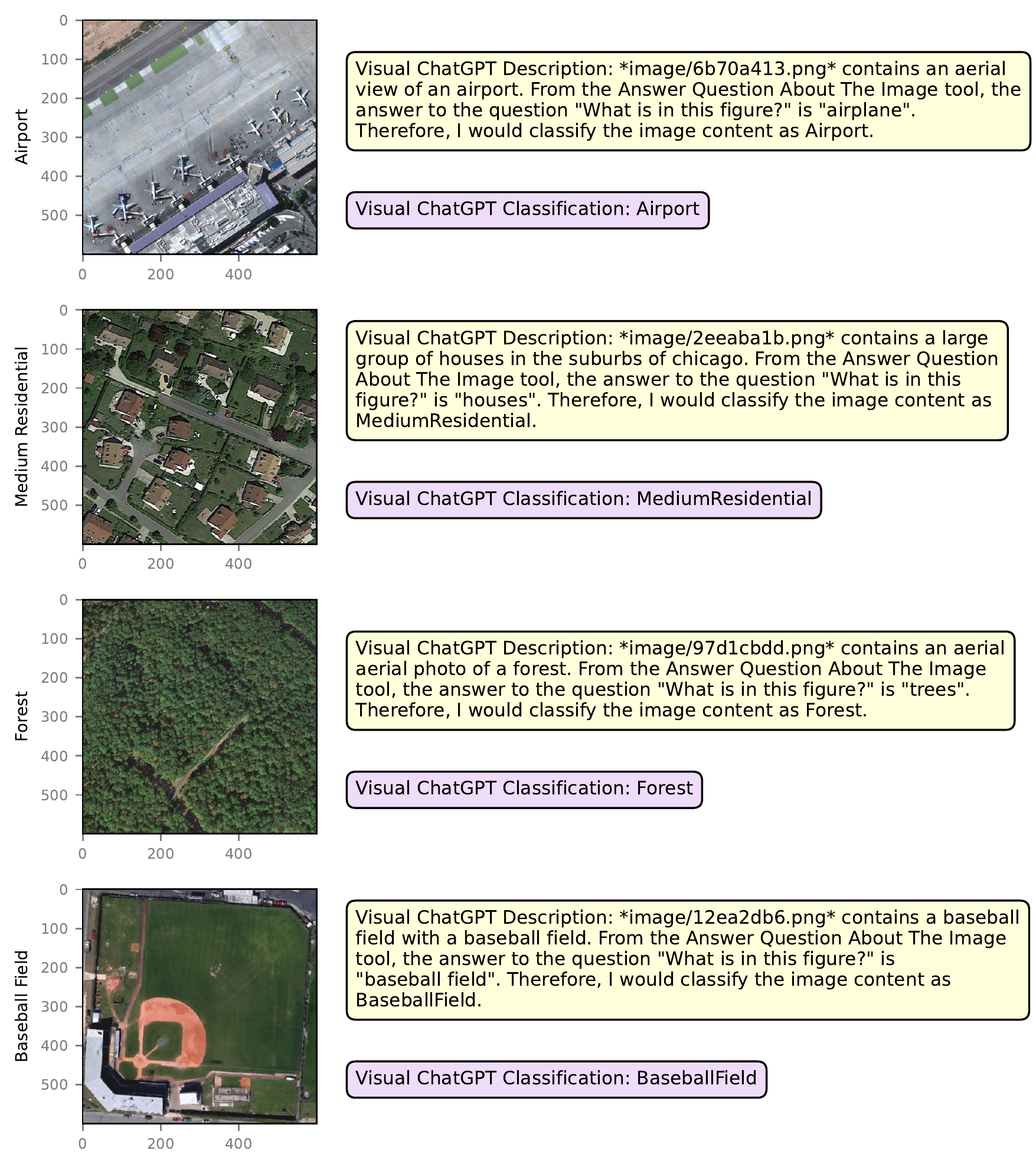}
\caption{\small \centering Sample images with correct Visual ChatGPT descriptions and classifications. For each image, two accompanying text boxes were provided. The first text box contains the description generated by Visual ChatGPT, while the second text box specifies the scene classification provided by the model. The images are arranged with each image being accompanied by a title on the left side, indicating its ground-truth label.\label{fig_scene3}}
\end{figure*}

\begin{figure*}[h!]
\centering
\includegraphics[width=17cm]{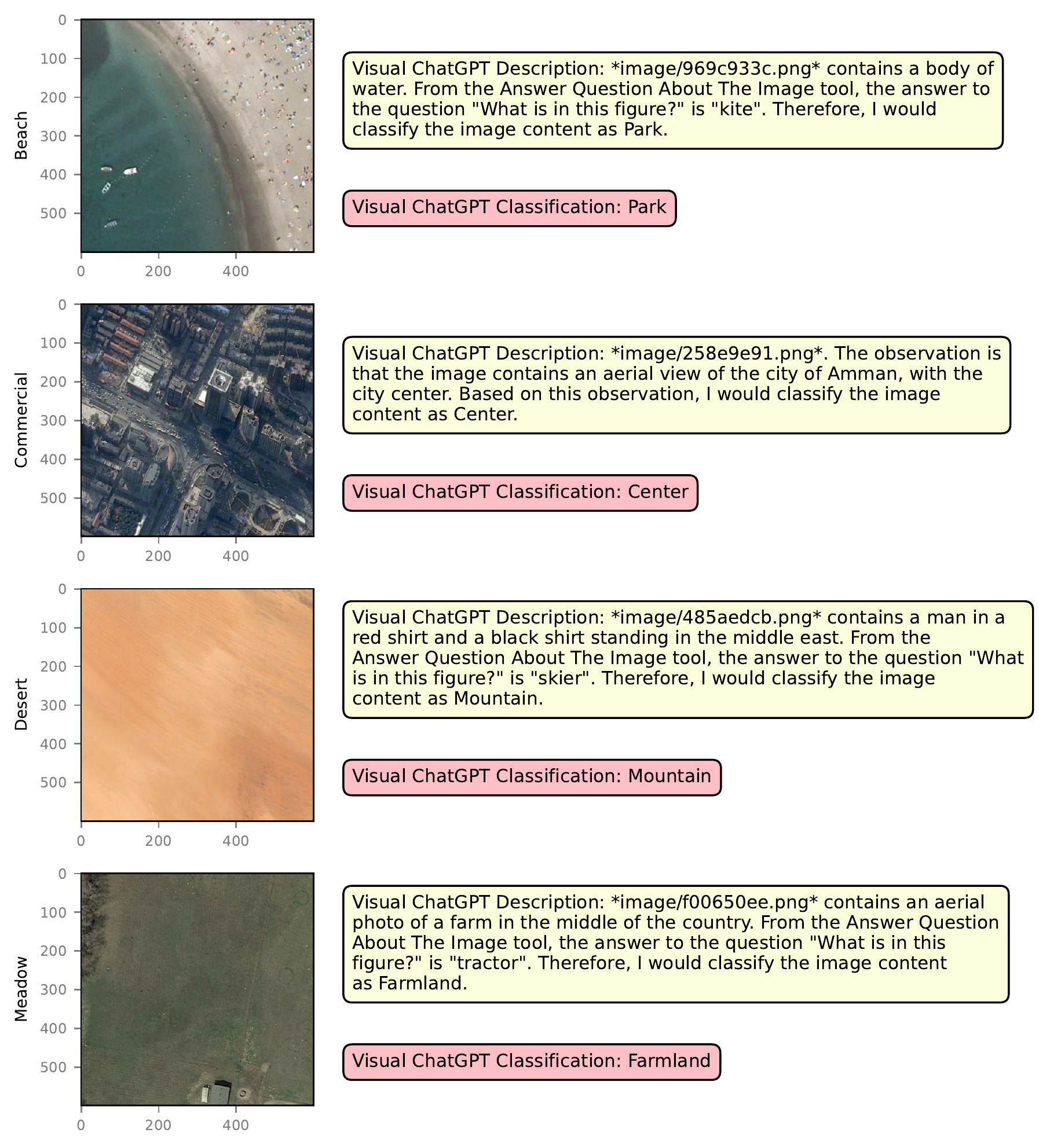}
\caption{\small \centering Sample images with incorrect Visual ChatGPT descriptions and misclassifications. Each image has a title specifying the true label of the scene, while the textboxes with incorrect descriptions and classifications are placed on the right side of each image.\label{fig_scene4}}
\end{figure*}

Figure~\ref{fig_scene3} showcases examples of instances that were accurately classified by the model.  Contrarily, Figure~\ref{fig_scene4} displays examples of instances inaccurately classified by it, demonstrating the necessity for additional tuning. Ensuring the incorporation of appropriate training sets into the learning process may further enhance the model's capabilities.

In the first example of Figure~\ref{fig_scene3}, an Airport, the model correctly identified the image as an aerial view of an airport with visible airplanes. The Medium Residential image example showcases the model's ability to detect a large group of houses. However, it incorrectly stated that these houses were located in the "suburbs of Chicago." The Forest scene example was also accurately classified, as the model identified it as an aerial photo of a forest with trees covering the landscape. Another instance, a Baseball Field scene, received a precise description as a baseball field with clear markings and layout. This was also the best-identified class in our tests.

The Visual ChatGPT model, however, misinterpreted and misclassified images across various classes, thus the reason why it presented lower accuracy overall. This highlights the challenges the model faces when handling aerial or satellite imagery, but it's mostly because it hasn't incorporate appropriate training sets of remote sensing data into its learning process.

The first example of Figure~\ref{fig_scene4} features a Beach, and the model recognizes the presence of a body of water and a "kite flying in the sky". However, Visual ChatGPT incorrectly classifies the image content as Park. This misclassification may have resulted from the additional objects present in the image. The Commercial example depicts an aerial view of a city center with various buildings, but Visual ChatGPT mistakenly classifies the image content as Center. This instance highlights the challenges in accurately classifying this dataset, primarily due to the similarities between urban centers and commercial areas. The Desert example showcases a desert landscape, but the model incorrectly assumes it contains "a person wearing a red shirt and black shorts in the Middle East". Oddly, Visual ChatGPT misclassifies the image content as Mountain. In the Meadow example, the model identifies the scene as an aerial photo of farmland, wrongfully noting a "visible tractor", and therefore erroneously classifies it as Farmland. 

The possible reasons for these mistakes can be attributed to the presence of similar features between the misclassified and true classes, or the model's reliance on specific visual cues that might not be present in every instance. These examples demonstrate the challenges and pitfalls in classifying certain aspects of an image. Nevertheless, some of the responses of Visual ChatGPT indicate its potential to accurately identify elements within these images, if fine-tuning and additional data training implementations were to be incorporated.

\subsection{Edge Detection}

In this section, we examine the performance of Visual ChatGPT's submodel in edge detection for remote sensing images. As the LoveDa dataset \cite{wang2022loveda} did not provide edge ground-truth labels created by human specialists, and considering the labor-intensive and challenging nature of the edge labeling task for innumerous objects, we opt to compare Visual ChatGPT's edge detection capabilities with the Canny and Sobel filters. This comparison highlights the similarities between the automated edge detection by Visual ChatGPT and these well-established methods.

The Canny edge detection method is generally more accurate and robust to noise compared to the Sobel edge detection. It is particularly useful for remote sensing images, where the presence of noise is common due to atmospheric effects, sensor limitations, or image acquisition conditions. The filter is effective in detecting continuous edges and suppressing noise, which is essential for accurately delineating features and boundaries in the images. 

The Sobel edge detection algorithm is computationally efficient, making it suitable for large-scale remote sensing data processing. However, the Sobel edge detection method is more susceptible to noise compared to the Canny edge detection, which might lead to false edges or missing features. Despite its limitations, Sobel edge detection can still provide valuable information about the presence and direction of edges, particularly when applied to high-quality remote sensing images with minimal noise.

\begin{figure*}[h!]
\centering
\includegraphics[width=\textwidth]{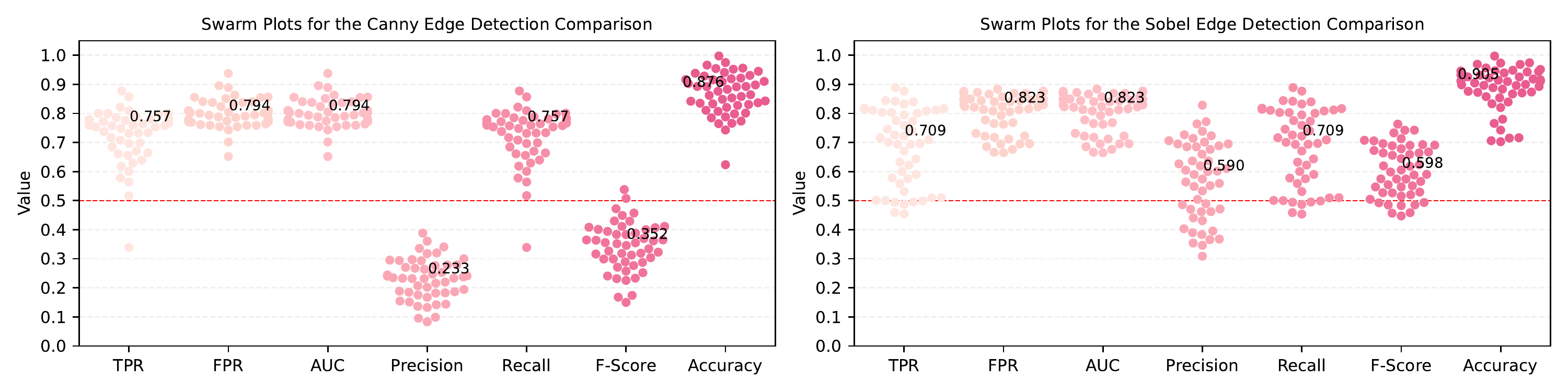}
\caption{\small \centering Swarm comparison of the performance metrics for both Canny and Sobel edge detections. The swarm plot displays the distribution of values measured by the multiple pairs of compared images, with the median value labeled. Although not all individual data points are shown, the swarm plot gives a general indication of the trend of the values. We included a red dashed line at y=0.5 to indicate the "random-guess" point.\label{fig_edge1}}
\end{figure*}

Figure~\ref{fig_edge1} illustrates that, for most image pairs, Visual ChatGPT achieves a True Positive Rate (TPR) above the "random-guess" threshold. However, due to the high False Positive Rate (FPR) observed, its Precision and F-Score are understandably lower than the other metrics.

When examining the TPR values, the edge detector model employed by Visual ChatGPT, which is based on the Canny edge from the OpenCV library, demonstrated greater similarity to our Canny edge filter compared to the Sobel filter. This outcome aligns with expectations since they are based on the same method, but considering we manually adjusted the Canny filter parameters to possibly yield superior visual results for each image. The findings are noteworthy as they reveal that the automated task performed by Visual ChatGPT closely approximates what a human might deem suitable.

\begin{figure*}[h!]
\includegraphics[width=\textwidth]{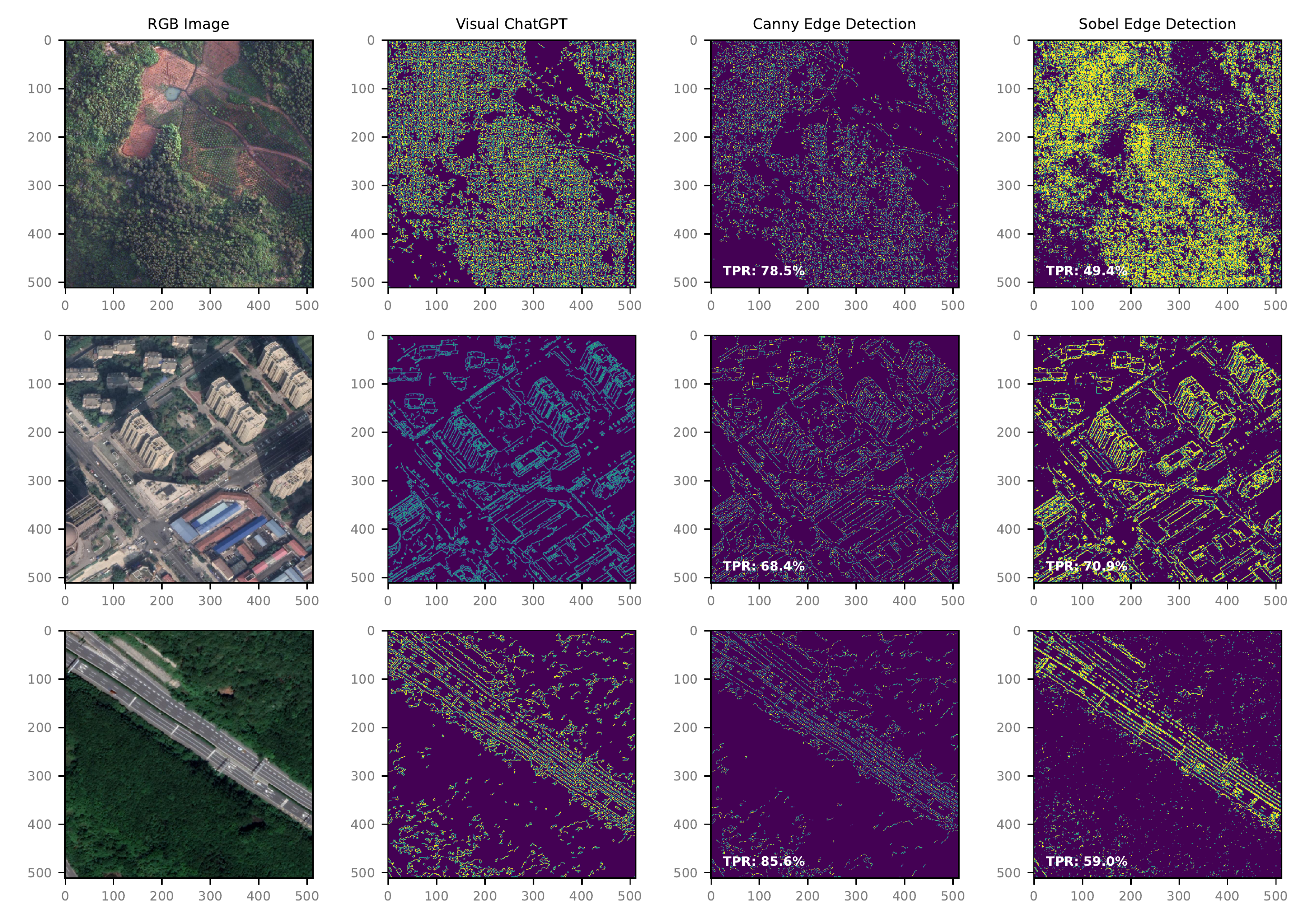}
\caption{\small \centering A comparison of the edge detection techniques on three example images. The visualizations are displayed using the “viridis” colormap symbolizing the magnitude of the detection, specifically in Sobel’s. The TPR values of the Canny and Sobel images in comparison to Visual ChatGPT’s detection are overlaid in the lower-left corner.\label{fig_edge2}}
\end{figure*}

However, it is crucial to acknowledge the substantial FPR and the low F-Score values. This can be primarily attributed to Visual ChatGPT's detector being sensitive to certain types of land cover, particularly in densely forested areas and heavily populated urban regions. Figure~\ref{fig_edge2} presents image examples of the detection results in such locations, which exhibit overall enhanced similarity with both Canny and Sobel filters.

\begin{figure*}[h!]
\includegraphics[width=\textwidth]{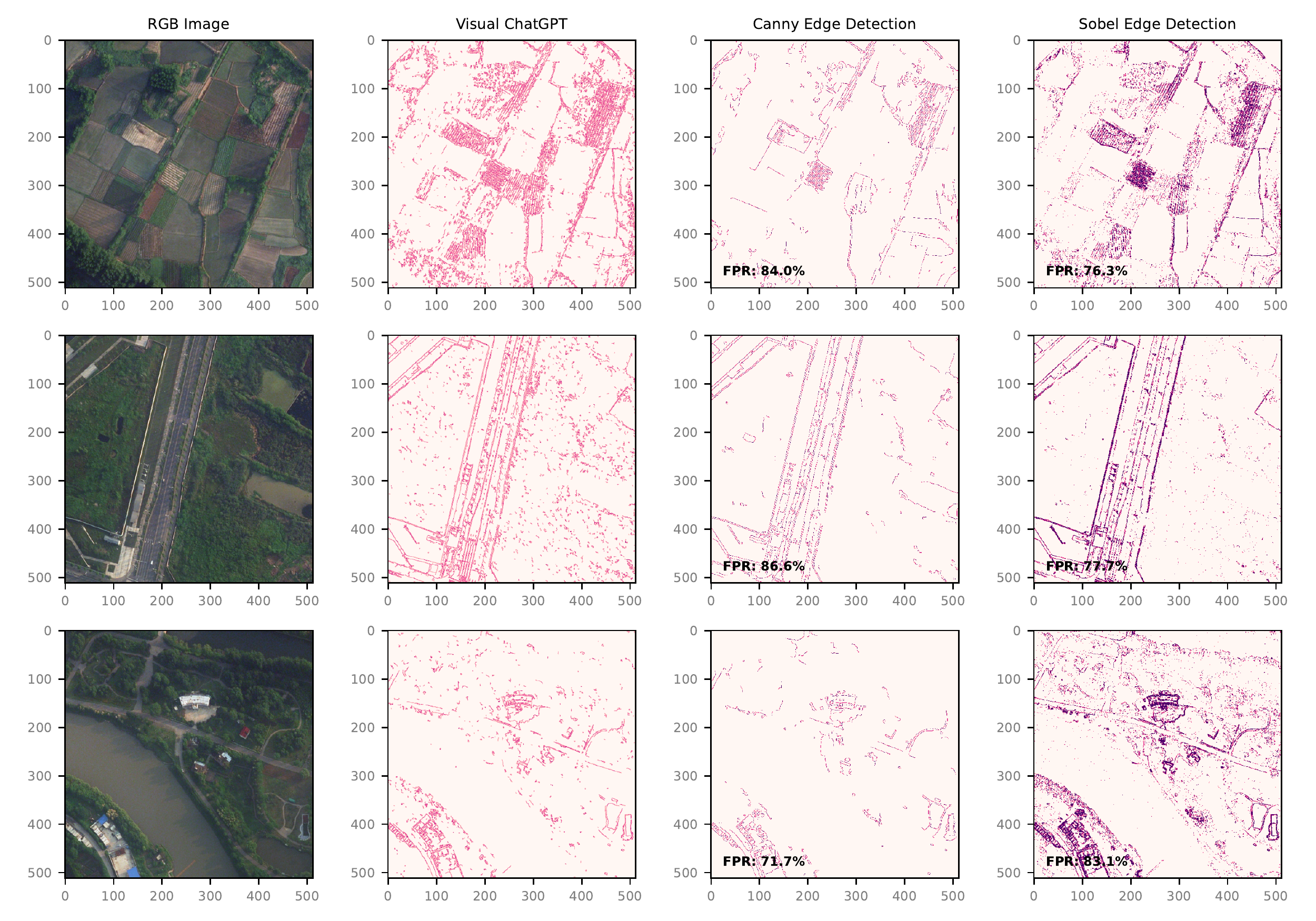}
\caption{\small \centering A visual comparison of edge detection techniques applied to three example images that returned low similarity. The visualizations use the 'RdPu' colormap indicating the magnitude of the edges, specifically useful for visualizing Sobel’s detection. The FPR values, comparing both images with Visual ChatGPT’s result, are displayed in the lower-left corner of the respective Canny and Sobel images.\label{fig_edge3}}
\end{figure*}

In areas covered with vegetation, Visual ChatGPT exhibited greater sensitivity than the Canny filter, though not as much as the Sobel filter. This pattern was also observed in built-up regions, particularly those with taller structures. Despite these limitations, Visual ChatGPT is capable of providing visually pleasing results in specific instances, such as detecting roads and bodies of water edges. However, the model generated a significant number of False Positives, which is undesirable as it introduces noise when interpreting the image. Figure~\ref{fig_edge3} showcases image examples where the FPR was among the highest observed, illustrating how farmlands and even less dense vegetation can influence the detection process.

These images demonstrate the differences in edge detection performance between the Canny and Sobel methods, as they indicate how difficult it is to extract this feature in certain conditions or areas characteristics. To enhance Visual ChatGPT's edge detection model on such instances, it is crucial to fine-tune it using a dataset tailored for edge detection tasks, incorporating proven methods like the Canny or Sobel filters, and adopting regularization techniques to prevent overfitting. Additionally, augmenting training data, evaluating alternative architectures, utilizing ensemble methods, and applying post-processing techniques can also further improve the model's performance. By adopting these strategies, Visual ChatGPT could deliver more accurate and reliable edge detection results.

\subsection{Straight Line Detection}

\begin{figure*}[h!]
\centering
\includegraphics[width=11cm]{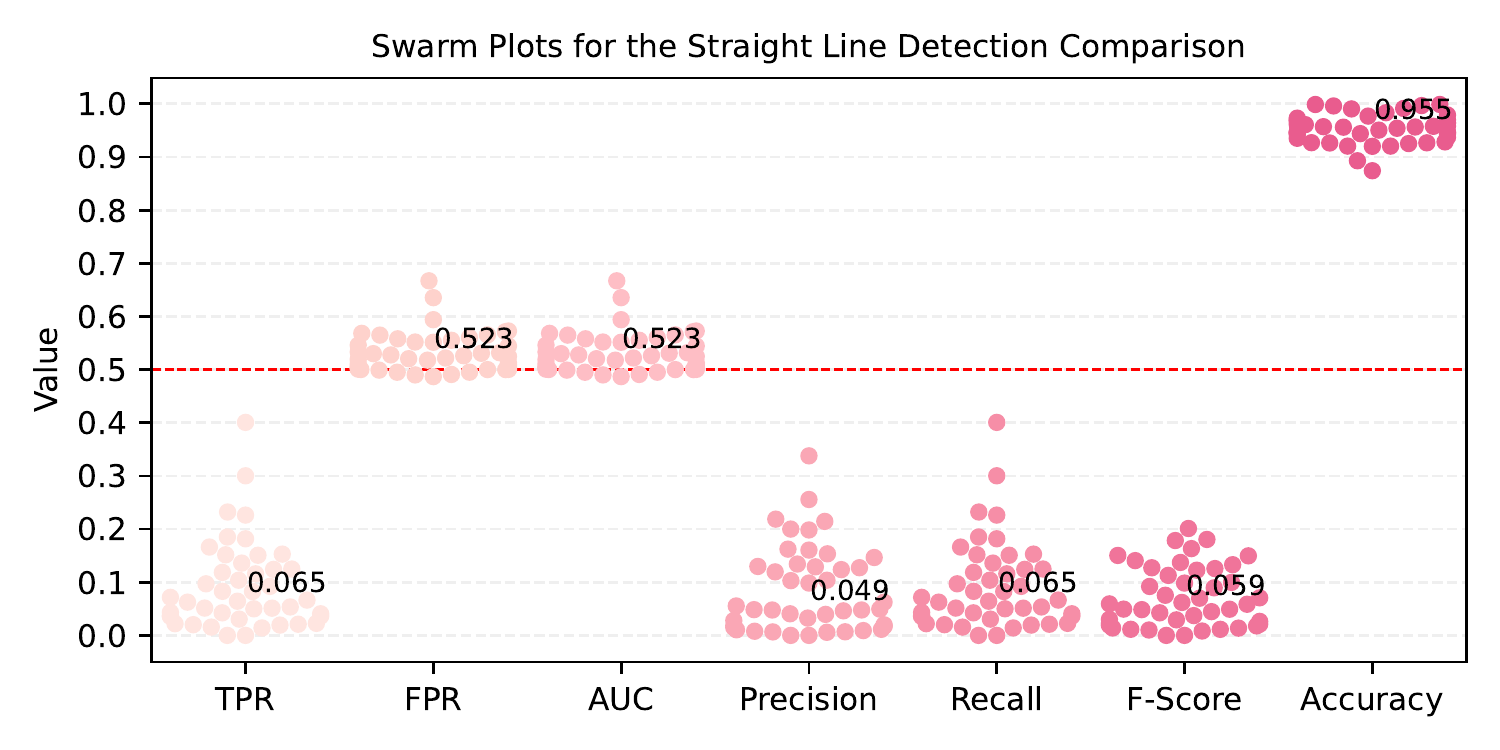}
\caption{\small \centering A swarm plot comparing performance metrics for the straight line detection model from Visual ChatGPT. The plot displays the distribution of values for each metric, with median values indicated in black text. We include a red dashed line at y=0.5 as a reference point for the "random-guess" threshold. While not all individual data points are displayed, the swarm plot provides an overall representation of the direction of the values.\label{fig_line1}}
\end{figure*}

Straight line detection in remote sensing images serves various purposes, such as building extraction, road detection, pipeline identification, etc. It proves to be a potent tool for image analysis, offering valuable insights for users. The evaluation of Visual ChatGPT's model for detecting straight lines employed the same protocol as edge detection. However, unlike the previous approach, we used manually labeled images, providing a more accurate ground-truth sample. Figure~\ref{fig_line1} presents a swarm plot illustrating the evaluation metrics used to compare Visual ChatGPT's detection results with their respective ground-truth counterparts.

\begin{figure*}[h!]
\includegraphics[width=\textwidth]{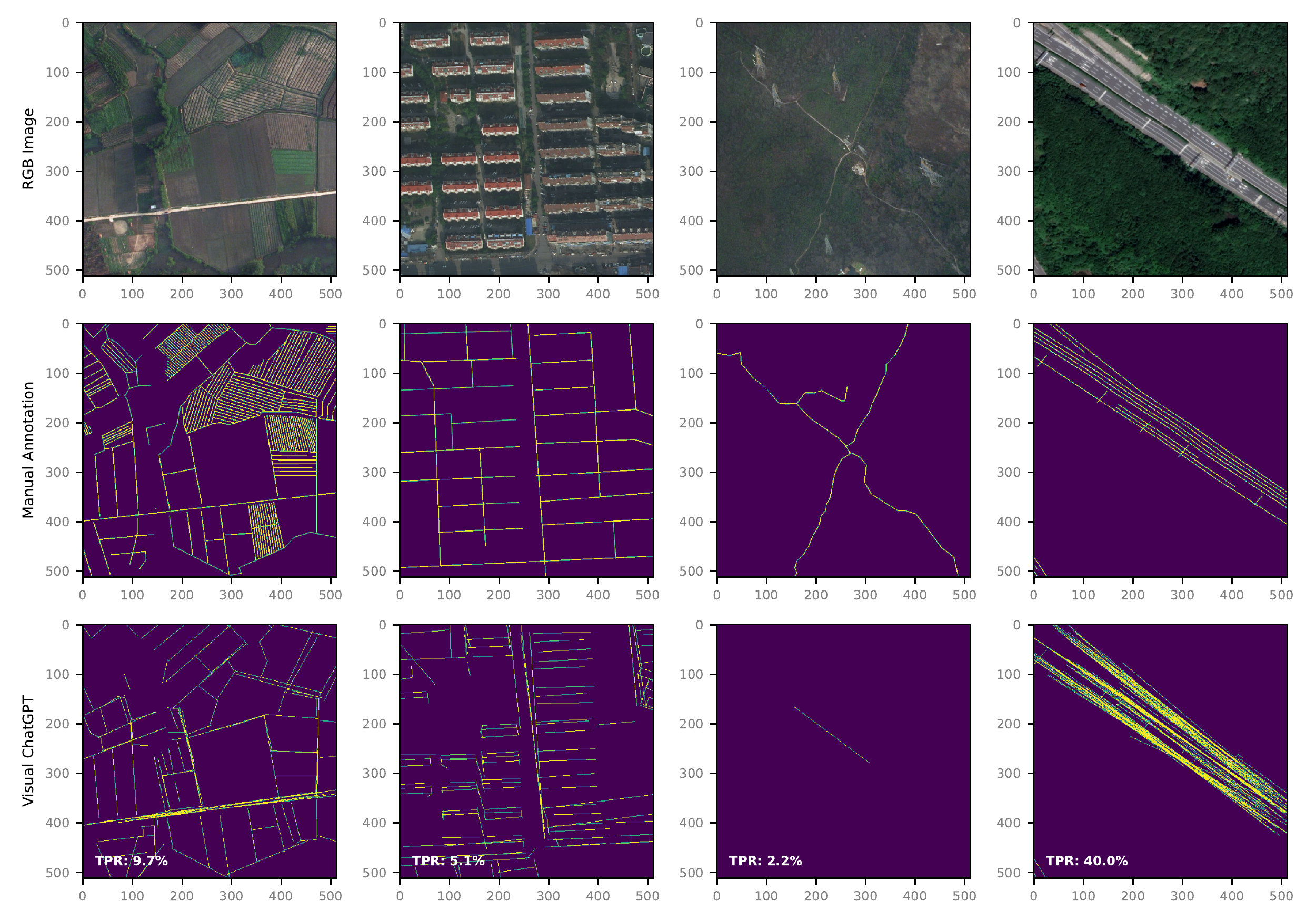}
\caption{\small \centering Comparative visualization of original RGB images (top row), manually annotated images (middle row), and Visual ChatGPT-generated images (bottom row) for four different sets. True positive rate (TPR) values are displayed in white text on the ChatGPT-generated images.\label{fig_line2}}
\end{figure*}

The results revealed that, concerning line detection, Visual ChatGPT's performance was quantitatively subpar. Given that lines typically constitute a small proportion of an image's pixels, metrics such as Accuracy are not well-suited for accurate measurement due to significant class imbalance. Moreover, the model generated a strikingly high number of False Positives compared to its TPR, primarily because it identified certain object edges as lines. To address this issue and provide a clearer understanding, we showcase image examples in Figure~\ref{fig_line2}, which highlight the disparities in line detection between rural and urban areas. By examining such visual comparisons, we noted the model's limitations and potential areas for improvement.

As observed, farmland areas exhibit a large number of lines, primarily due to plantations and tractor roads between them. Identifying these lines can be challenging, even for human specialists. However, Visual ChatGPT managed to detect a considerable number of roads interspersed among the plantation fields. It was capable of identifying the boundaries of these fields, which is an important aspect of feature extraction for these areas. In urban settings, however, extracting streets can be difficult, mainly because objects and shadows partially obscure them. These are also heavily dense areas, with multiple objects overlapping the streets.

Figure~\ref{fig_line2} also highlights the overall best and worst results in its 3rd and 4th columns, featuring dirt roads and a paved highway, respectively. For the dirt roads, it is understandable that their winding nature may pose a challenge for the model. Conversely, the paved highways represent the best overall detections by Visual ChatGPT, showcasing its potential in these contexts.

Improving Visual ChatGPT's line detection and extraction capabilities in remote sensing imagery involves practically the same procedures as described previously, like fine-tuning the model on a tailored dataset, augmenting training data, and also applying pre-processing techniques to enhance input image quality. Additionally, incorporating domain-specific knowledge, exploring alternative model architectures, utilizing ensemble methods, and employing enhanced post-processing techniques can further optimize its performance on returning satisfying results.

\subsection{Image Segmentation}

As stated, image segmentation is the process of partitioning an image into homogeneous regions based on features such as color, texture, or spectral properties, with multiple applications in image analysis. However, for the Visual ChatGPT model, handling remote sensing data can be challenging due to the diverse and complex nature of these images. Factors such as varying spatial resolutions, the presence of shadows, seasonal variations, and spectral similarities among different land cover types may hinder the model's performance, necessitating further optimization or the integration of domain-specific knowledge to effectively address these complexities. Still, VLMs can provide a valuable approach to the image segmentation task by enabling non-expert users to perform segmentation using text-based guidance. This capability has the potential to be integrated into remote sensing applications. 

However, in the case of Visual ChatGPT, our tests with various prompts revealed that controlling the "Segmentation on Image" tool was not as feasible as it was for the "Get Image Description" and "Answer Question About Image" tools. Consequently, we were unable to guide Visual ChatGPT to segment specific classes from our images. As a reminder, since classification metrics like Precision, Recall, and F-Score necessitate matching classes in both ground-truth and predicted values, these metrics were unsuitable for comparing Visual ChatGPT's performance in this task. Instead, we employed metrics that assessed the similarity between image pairs, which, when combined with qualitative analysis, offered insight into the model's effectiveness in handling this type of data.

\begin{figure*}[h!]
\centering
\includegraphics[width=11.5cm]{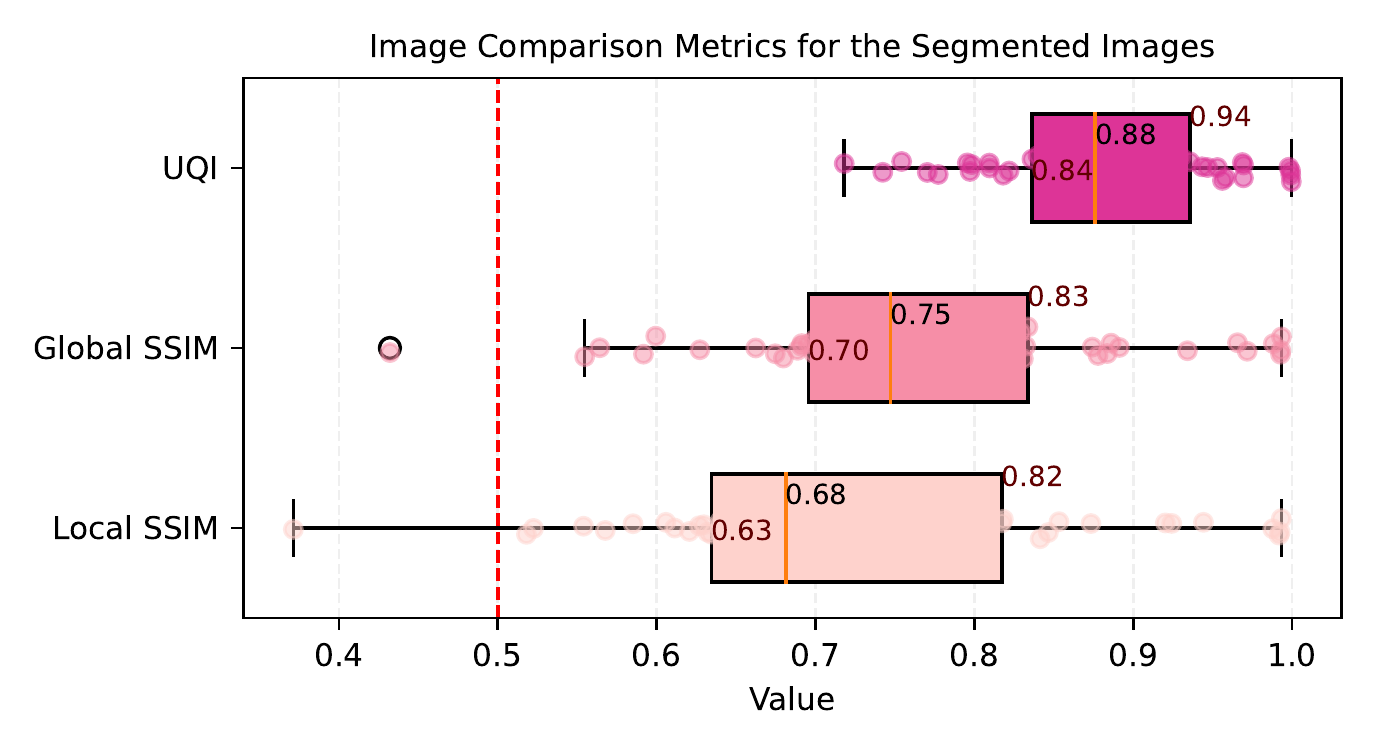}
\caption{\small \centering Horizontal box plots comparing image comparison metrics (Local SSIM, Global SSIM, and UQI) for the segmented images with the Visual ChatGPT model. The 25th, 50th (median), and 75th percentiles are displayed on each box plot, allowing for a clear assessment of the central tendency and spread of the data, and a red dashed line at x=0.5 serves as a reference point.\label{fig_segmentation1}}
\end{figure*}

To evaluate the predictions of Visual ChatGPT, we compared the ground-truth data from the LoveDA dataset \cite{wang2022loveda} to the segmented images generated by the model. Figure~\ref{fig_segmentation1} presents the values of both Local and Global SSIM metrics, as well as the UQI values for this comparison. The Local SSIM metric is particularly noteworthy in this context, as it is designed to focus on local variations during image analysis. Meanwhile, the Global SSIM calculates a score for the entire image, offering a comprehensive assessment of overall similarity. The UQI metric compares structural information based on luminance and contrast between colors, making it a more suitable metric for overall performance.

\begin{figure*}[h!]
\includegraphics[width=\textwidth]{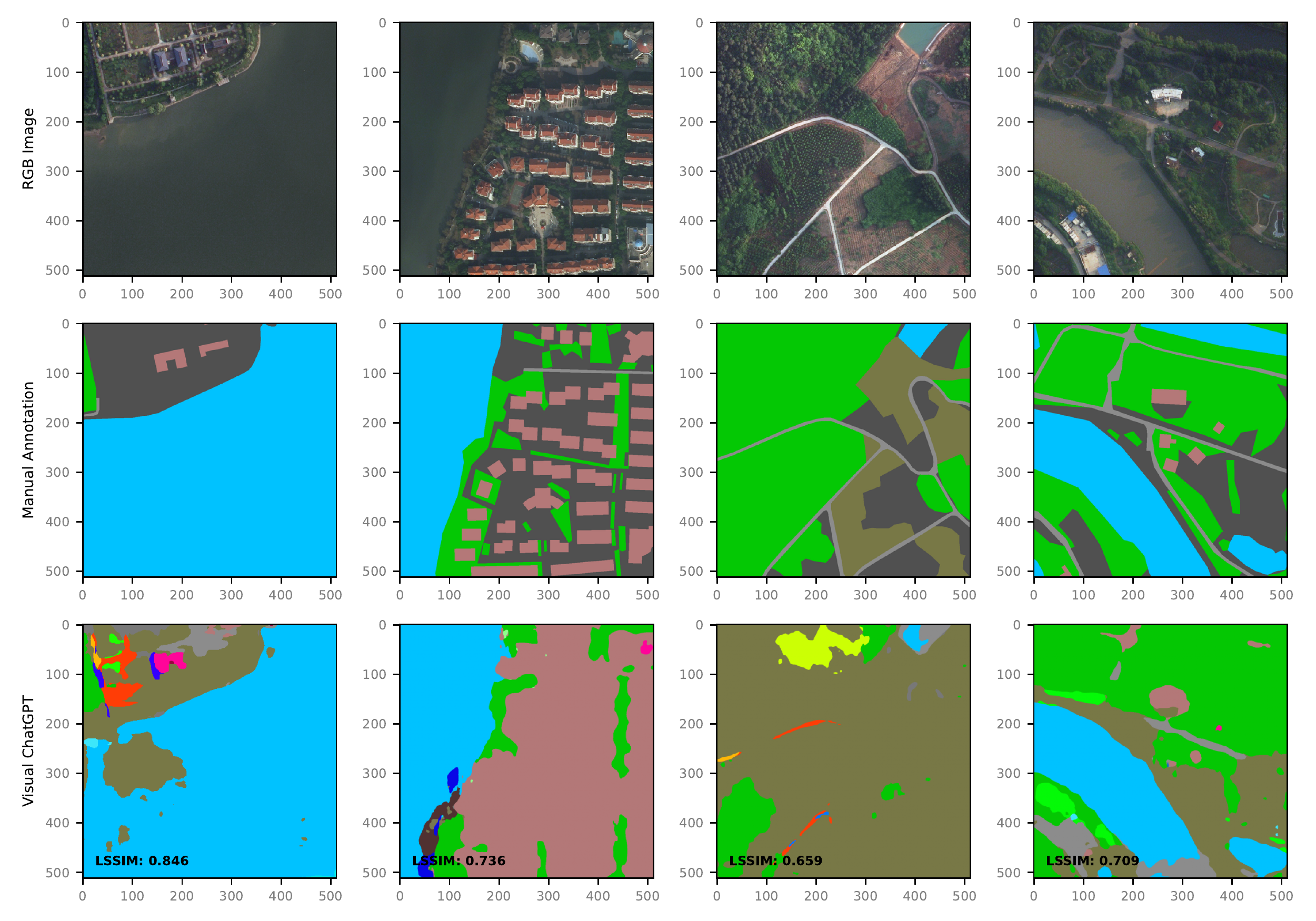}
\caption{\small \centering Examples of labeled images compared to the Visual ChatGPT segmentations that scored higher on the similarity metrics. In the bottom row, Local SSIM (LSSIM) values are displayed in the left corner of each segmented image, providing a quantitative measure of the similarity between the annotations and the Visual ChatGPT segmentations.\label{fig_segmentation2}}
\end{figure*}

\begin{figure*}[h!]
\includegraphics[width=\textwidth]{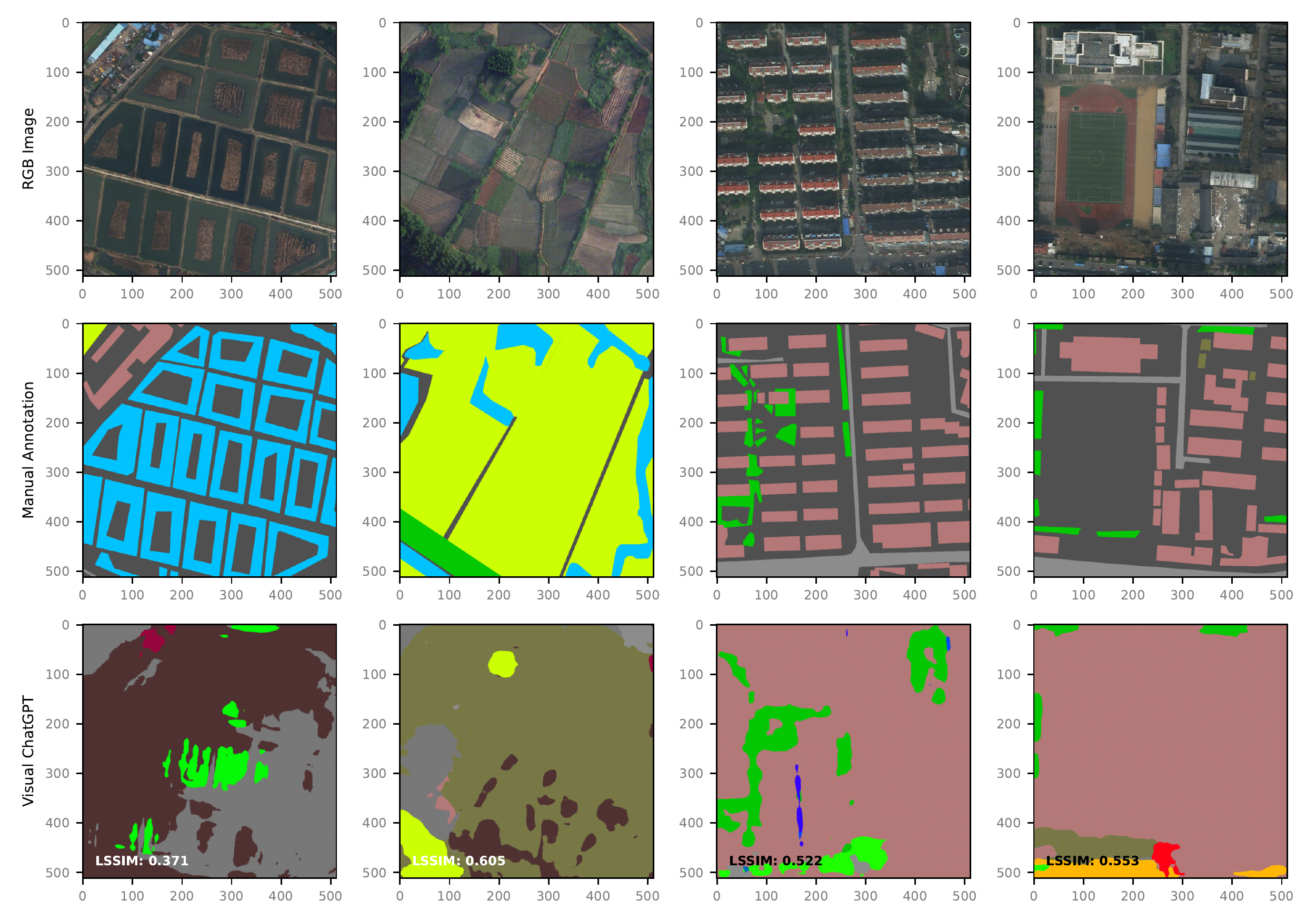}
\caption{\small \centering Examples of labeled images juxtaposed with Visual ChatGPT segmentations that scored the lowest on similarity metrics. In the bottom row, LSSIM values are shown, in black or white depending on its background, for each segmented image, offering a quantitative assessment of the dissimilarity between the ground-truth and the model’s segmentations.\label{fig_segmentation3}}
\end{figure*}

In our comparison, the majority of the data revealed notable similarity values, with more pronounced negative effects on local analysis (Local SSIM) than on the full-scale (Global SSIM and UQI) assessment. These images predominantly featured farmlands, as well as scenes with both urban and rural elements, resulting in a more varied landscape. Contrarily, some images exhibited high similarity with the ground-truth data. These images typically displayed less diverse features, such as extensive vegetation cover, large bodies of water, or densely clustered structures of a similar nature. To corroborate this, Figures~\ref{fig_segmentation2} and~\ref{fig_segmentation3} were included, showcasing both the challenges and potential of the Visual ChatGPT segmentation model. This visual comparison enables a clear evaluation of the model's performance to the manual annotations.

Visual ChatGPT utilizes a powerful image segmentation model underneath, thus making it an impressive tool. However, its knowledge is not specifically associated with aerial or satellite imagery, but more with the terrestrial type of images, while the segmentation classes are more diverse. Additionally, the model was not effective in incorporating additional textual information to segment remote sensing images, as our tests have shown that by asking the model to segment images, with or without human instructions, it yielded the same results. Furthermore, Visual ChatGPT did not indicate appropriately which classes it has segmented over the investigated images, even when prompted with a specific command. Instead, the model segments the image and uses the "Answer Question about Image" function to respond to it, using information about the context of the original RGB image rather than the labels/classes that it identified.

The segmentation model demonstrates both potential and challenges when dealing with various land cover types. While the model shows promising performance in images with less diverse features or densely clustered structures of a similar nature, it encounters difficulties in accurately segmenting more complex scenes. The difficulties primarily arise in the local analysis, as evidenced by lower Local SSIM values, which could be attributed to the model's limited exposure to such diverse data during training.

Nonetheless, Visual ChatGPT's ability to achieve high similarity with ground-truth data in certain cases indicates that, with targeted improvements, it could be adapted to effectively handle a wider range of land covers and deliver more accurate segmentation results. As such, to fully realize the potential of Visual ChatGPT in these scenarios, further improvements and fine-tuning are required to better handle the diverse and intricate characteristics of different land types.

\section{Discussion}

The investigation into the Visual ChatGPT model's proficiency in handling remote sensing imagery yielded intriguing results, indicating both its potential and limitations. While the overall model accuracy of 38.1\% is considerably higher than the random chance baseline of 5.88\% in a 17-class classification task, there were notable disparities in performance across different classes. The model exhibited proficiency in accurately identifying scenes containing Baseball Fields, Bridges, Beaches, and Mountains, as demonstrated by high F-Scores. However, it faced challenges recognizing and classifying Bareland, Meadows, and Deserts, evidenced by lower performance metrics. Additionally, the model encountered difficulties distinguishing urban scenes such as Commercial, Church, Center, Industrial, and Dense Residential areas.

The edge detection analysis revealed that the model demonstrated similarity to our adjusted Canny edge filter. Despite the similarity, the substantial False Positive Rate and the low F-Scores, particularly in densely forested areas and heavily populated urban regions, highlight a crucial area for improvement. The model's performance in straight-line detection was also mixed. It demonstrated potential in farmland areas by detecting numerous roads interspersed among plantation fields and boundaries of fields. However, it struggled with the extraction of streets in urban settings and the winding nature of dirt roads. Conversely, it performed optimally when detecting paved highways, which suggests a solid foundation on which future optimizations can be built.

Image segmentation is another area that highlighted the model's potential and its current limitations. The ViT implemented model demonstrated strong performance in images with less diverse features or densely clustered structures of a similar nature but faced difficulties accurately segmenting more complex scenes. It also didn't effectively leverage additional textual information to improve segmentation results, a feature that would be a significant enhancement to be implemented in future versions of it. While it is evident that the model can correctly interpret and classify images across several classes, it also made mistakes, underlining the importance of further model fine-tuning and incorporation of more diverse and representative training datasets.

As stated, the "Segment on Image" function incorporates the Uniform model \cite{li2022uniformer}, a vision-based transformer that was not specifically designed for remote sensing data. While not specifically trained for it, its architecture enables it to reduce local redundancy and capture global dependency effectively, which could be the reason behind the segmentation results in some cases. As such, it was capable of segmenting a broad range of land covers, although not without its mistakes. The recent literature, however, suggest that models based on ViT can be capable of performing zero-shot segmentation on different domains, or at least be adapted with few-shot learning \cite{Sun2021, kirillov2023segment, zhang2023personalize, qiusheng_wu_2023_7966658}.

ViT-based models currently represent the state-of-the-art in handling remote sensing data as they have triumphed in areas where traditional Convolutional Neural Networks (CNNs) faced challenges. The potential of these models has already been demonstrated, but only when specifically trained with remote sensing data \cite{Aleissaee2023}. In different land cover segmentation and classification tasks, models such as SegFormer, UNetFormer, and RSSFormer returned impressive results, with F-Scores values above 90\% \cite{Wang2022, Gonalves2023, Xu2023}. Furthermore, since the current segmentation model is not capable of discerning text-to-image, an integration with capable LLMs with the ViT models may improve the segmentation of these images \cite{zhang2023visionlanguage}.

As last, in the current state of its development, Visual ChatGPT may present certain challenges for non-experts in the realm of image processing tasks. The complexity of the interface and operations, an inherent characteristic of this early-stage technology, poses a potential barrier to its widespread adoption. Our research delineates the significant potential of Visual ChatGPT for remote sensing tasks; however, the transition from potential to practical usage necessitates further improvements, primarily targeted at enhancing its user-friendliness. We envisage that the near future will witness concerted efforts towards improving the usability of such models, fostering an environment conducive for both experts and non-experts. We anticipate these improvements to manifest in the form of more intuitive user interfaces and comprehensive guidance, thus broadening the accessibility and usability of Visual ChatGPT.

\section{Improving Visual Language Models for Remote Sensing Analysis}

In this section, we provide a  broader vision of Visual Language Models (VLMs) in remote sensing analysis and discuss possibilities for future implementations. While our experiments focused on Visual ChatGPT, it is clear that novel VLMs will be able to tackle different tasks and be useful, in general, in multiple domains. VLMs are a class of machine learning models that are designed to understand and generate content that combines both visual and textual information \cite{mialon2023augmented}. VLMs are trained to associate images with their related text, and this enables them to carry out tasks that involve understanding and generating such multimodal content \cite{alayrac2022flamingo}. VLMs are often built by combining techniques from the fields of computer vision, which focuses on understanding and processing images, and NLP, which focuses on understanding and processing text. As Visual ChatGPT is one of the many VLMs that are surging recently, it is important to discuss their involvement with image manipulation and how they can be adapted into the remote sensing domain.

With the constantly increasing amount of remote sensing data available, there is a growing need for efficient methods to process and analyze this data \cite{Chi2016}. As VLMs continue to evolve and improve, their applications in multiple fields are expected to expand significantly. By incorporating additional techniques and algorithms, it can become a powerful tool for non-experts to analyze and understand complex remote-sensing images. In this section, we explore the future perspectives of these technologies in remote sensing practice, discuss possible applications, and outline the necessary research directions to guide their development and improvement. 

Firstly, to apply VLMs to remote sensing data, it would be necessary to collect a large dataset of labeled images. This may involve manually annotating the images, which can be a time-consuming and expensive process \cite{Sun2021}. Alternatively, transfer learning techniques can be used to fine-tune pre-trained models on a smaller set of labeled images, possibly reducing the amount of labeled data required for training \cite{Tong2020}. By learning from a limited number of examples, few-shot learning models, for instance, can develop better generalization capabilities \cite{alayrac2022flamingo}, as they can be more robust to variations in remote sensing data. Such an approach can enable the models to recognize and analyze unique features, patterns, and structures present in satellite or aerial images, thereby significantly improving their performance and applicability in this domain.

By adapting VLMs like Visual ChatGPT for remote sensing analysis, we can also create powerful tools to aid professionals, students, and enthusiasts in their work. These models can facilitate the development of image and data processing, provide guidance in choosing and applying the most appropriate algorithms and techniques, and offer insights into the interpretation of remote sensing data \cite{Lobry2020}. The models can help users overcome coding challenges, offer guidance on data processing techniques, and facilitate collaboration between individuals with varying levels of expertise and study fields \cite{zhang2023adding, zhang2023visionlanguage}. In turn, this assistance can enhance the efficiency and accuracy of remote sensing workflows, allowing them to focus on higher-level tasks and decision-making.

A potential for Visual ChatGPT or VLMs, in general, is that they can be seamlessly integrated with a variety of geospatial tools and platforms to significantly elevate user experience. By combining advanced models with existing geospatial software, toolboxes, or cloud-computation platforms, users can access an enriched suite of functionalities that cater to a wide range of applications. This integration not only amplifies the capabilities of existing tools \cite{mialon2023augmented} but also unlocks innovative possibilities for analyzing and interpreting geospatial data. By leveraging the natural language understanding and visual processing abilities of VLMs, the interaction with these platforms can become more intuitive, leading to improved efficiency and accessibility.

In essence, the improved versions of VLMs can be applied to a wide range of remote sensing tasks. These applications can benefit from the model's ability to provide real-time feedback, generate code snippets, and analyze imagery, thus streamlining the overall process. For example, a model could be trained to identify common patterns in remote sensing data and generate code to automatically detect and analyze these patterns. This has the potential to help to speed up the processing of large datasets and minimize the intricacies of manual intervention.

As for applications, VLMs can be expanded to encompass various essential image tasks, such as texture analysis, principal components analysis, object detection, and counting, but also curated to domain-specific remote sensing practices as well. By integrating change detection algorithms \cite{Shafique2022} into these VLMs, for instance, users can interact with the models to automatically identify landscape alterations, facilitating the monitoring and assessment of the impacts caused by human activities and natural processes on the environment. Anomaly detection, a technique that identifies unexpected or unusual features in remote sensing images \cite{Hu2022}, can also greatly benefit from this integration. Time series analysis is also a valuable method that involves analyzing changes to reveal patterns, trends, and relationships in land cover \cite{Gmez2016} and could be added to it. Consequently, by incorporating tailored algorithms into VLMs, users can examine multiple images over time, gaining insights into the dynamics of the Earth's surface.

Furthermore, the integration of machine and deep learning algorithms specifically designed for remote sensing applications, such as convolutional neural networks and vision transformers \cite{Li2022, Aleissaee2023}, can help enhance the performance and capabilities of visual models. These methods can improve the VLM's ability to recognize and analyze complex patterns, structures, and features in remote sensing images, leading to more accurate and reliable results. Currently, there are multiple networks and deep learning models trained for various remote sensing tasks that are available and could be potentially implemented \cite{Bai2022, Papoutsis2023}.

Overall, the potential for VLMs like Visual ChatGPT to aid in remote sensing image processing is vast and varied. As the technology continues to evolve and improve, we will likely see an increasing number of innovative applications in this field, with new features and capabilities being developed to meet the specific needs of users. Looking to the future, it is likely that VLMs will continue to play an increasingly important role in image data analysis. As these models become more advanced and better integrated with existing tools and workflows, they have the potential to greatly improve the efficiency and accuracy of remote sensing practices.

Although our experiments with Visual ChatGPT only consist of one perspective, VLMs have, in general, an important role in image analysis. In short, to guide the development and improvement of VLMs in remote sensing, several research directions could be explored:
\begin{itemize}
    \item Investigating the  optimal methods and strategies for fine-tuning and adapting models to remote sensing tasks;
    \item Developing performance benchmarks and evaluation metrics specific to remote sensing applications on these models;
    \item Exploring the integration of these models with other remote sensing tools and platforms, such as Geographic Information Systems (GIS), for a seamless user experience;
    \item Conducting user studies to understand how the models can best work for these data and how they can be adjusted to user behavior;
    \item Studying the limitations and biases of the models when applied to remote sensing imagery, and devising strategies to mitigate them.
\end{itemize}

\noindent And, in terms of applicability, the following areas can also be considered to be pursued, thus contributing to enhancing the development of VLMs in remote sensing imagery processing:
\begin{itemize}
    \item Investigating the effectiveness of incorporating domain-specific knowledge and expertise into the models, such as spectral indices;
    \item Examining the scalability and efficiency of the models when working with large-scale remote sensing datasets;
    \item Assessing the robustness and generalizability of the models across various remote sensing data types, including multispectral, hyperspectral, Synthetic-Aperture Radar (SAR), and LiDAR;
    \item Evaluating these models for real-time or near-real-time remote sensing analysis;
    \item Exploring the potential of combining VLMs with other advanced machine learning techniques, such as reinforcement learning;
    \item Investigating the implementation for data fusion tasks, where information from different remote sensing sensors or platforms are combined.
\end{itemize}

\section{Conclusions}

In this study, we investigated the applicability and performance of Visual ChatGPT, a VLM, for remote sensing imagery processing tasks, highlighting its current capabilities, limitations, and future perspectives. We have demonstrated the effectiveness and problems of this model in various remote sensing tasks, such as image classification, edge and line detection, and image segmentation. Additionally, we have discussed its role in assisting users and facilitating the work of professionals, students, and enthusiasts in the remote sensing domain by providing an intuitive, easy-to-learn, and interactive approach to image processing.

In our investigation we found that, despite its ability to perform scene classification above the random-guess baseline, the model faced difficulties distinguishing certain landscape classes and urban scenes. The model showed potential in edge detection and straight-line identification, especially in farmland areas and on paved highways, but struggled in densely populated regions and complex landscapes. While the model's segmentation showed promising results in less diverse or densely clustered scenes, it faced difficulties in more complex environments. Still, although some results may not appear impressive, we believe that these initial findings lay a groundwork for future research and improvements.

While Visual ChatGPT shows promise in its current state, there is still plenty of room for improvement, fine-tuning, and adaptation to better suit the unique needs of remote sensing analysis. Future research could focus on optimizing the model by either fine-tuning with techniques such as few-shot learning, or improving their natural language capacities to recognize objects based on their class and segment them in a more guided manner, be it though label or text-based prompts. By doing so, we can unlock the capacity of these models in a wide range of remote sensing applications, varying from environmental monitoring and disaster management to precision agriculture and infrastructure planning.

In light of our findings, the integration of VLMs into remote sensing has immense potential to transform the way we process and analyze Earth's surface data. With continued evolution and adaptation to the specific needs of aerial/satellite data, these models can prove to be essential resources in assisting important challenges in image processing. It is crucial to emphasize the significance of ongoing research in this area and encourage further exploration of the capabilities of Visual ChatGPT, as well as other VLMs in dealing with remote sensing tasks in the near future.


\section*{Acknowledgements}

This study was financed in part by the Coordenação de Aperfeiçoamento de Pessoal de Nível Superior (CAPES) - Finance Code 001 and Print (88881.311850/2018-01). The authors are funded by the Support Foundation for the Development of Education, Science, and Technology of the State of Mato Grosso do Sul (FUNDECT; 71/009.436/2022) and the Brazilian National Council for Scientific and Technological Development (CNPq; 433783/2018-4, 310517/2020-6; 405997/2021-3; 308481/2022-4; 305296/2022-1).

\section*{Conflicts of Interest}

The authors declare no conflict of interest. The funders had no role in the design of the study; in the collection, analyses, or interpretation of data; in the writing of the manuscript, or in the decision to publish the results.

\section*{Abbreviations}{
The following abbreviations are used in this manuscript:\\

\noindent 
\begin{tabular}{@{}ll}
AI & Artificial Inteligence\\
AUC & Area Under the Curve\\
FN & False Negative\\
FP & False Positive\\
FPR & False Positive Rate\\
GIS & Geographic Information Systems\\
GPT & Generative Pre-trained Transformer\\
LLMs & Large Language Models\\
NLP & Natural Language Processing\\
SAR & Synthetic-Aperture Radar\\
SSIM & Structural Similarity Index Measure\\
TN & True Negative\\
TP & True Positive\\
TPR & True Positive Rate\\
UQI & Universal Image Quality Index\\
VLM & Visual Language Model
\end{tabular}
}

\normalsize


\end{document}